%% file: main.tex
\documentclass{article}

\usepackage{pifont}

 \usepackage[preprint]{neurips_2026}

\usepackage{xcolor}
\newcommand{\cmark}{\textcolor{green!60!black}{\ding{51}}}
\newcommand{\xmark}{\textcolor{red!70!black}{\ding{55}}}

\input{macros}

\author{
  Radu Beche\thanks{Work done while visiting the Astra-Vision group at Inria, Paris.} \\
  TUCN \\
  \And
  Raoul de Charette \\
  Inria \\
}
\begin{document}

\maketitle
\vspace{-1em}
\begin{center}
    \vspace{-1.5em}\url{https://rdbch.github.io/koalags}\\[1em]
\end{center}

\begin{abstract}

Deploying a trained 3D Gaussian Splatting scene under shrinking compute budgets requires one primitive set to remain effective when rendered at smaller prefixes.
In this continuous level of detail (LoD) setting, the challenge is to decide which Gaussians should retain capacity as the budget decreases. A pilot study on 3DGS-MCMC reconstructions shows that image complexity, computed from the input views before optimization, predicts where reconstruction error concentrates in image and 3D space.
It also reveals that simple regions contain too many Gaussians and degrade as much as complex ones at lower budgets, despite being inherently cheaper to represent. Subsequently, in KoalaGS, during training we use this complexity prior to bias relocation toward poorly reconstructed areas while ensuring smooth degradation across LoDs.
A primitive importance score then enables inference at low-capacity, while preserving the scene structure.
Tested across three datasets, KoalaGS maintains competitive full-capacity quality while significantly improving quality across the budget spectrum, yielding smoother transitions across LoDs, and enabling faster progressive streaming.

\end{abstract}

\input{sections/00_introduction}
\input{sections/01_related_work}
\input{sections/02_analysys_and_method}
\input{sections/03_results}
\input{sections/04_conclussion}

\clearpage
{
    \small
    \begin{ack}
This work was supported by the "Romanian Hub for Artificial Intelligence-HRIA", Smart Growth, Digitization and Financial Instruments Program, MySMIS no. 351416 and by the Astra Team, Inria. This work was granted access to the HPC resources of IDRIS under the allocation 2025-AD011017283 made by GENCI on the Jean-Zay Cluster. The authors would like to thank Sergiu Nedevschi and Fawzi Nashashibi for making this collaboration possible.

\end{ack}
    \bibliographystyle{unsrt}
    \bibliography{main}
}

\newpage

\newpage
\appendix
\raggedbottom
\input{sections/05_supp}

\end{document}

%% file: macros.tex
\usepackage[utf8]{inputenc}
\usepackage[T1]{fontenc}
\usepackage{hyperref}
\usepackage{url}
\usepackage{booktabs}
\usepackage{amsfonts}
\usepackage{nicefrac}
\usepackage{microtype}
\usepackage[table,xcdraw]{xcolor}
\usepackage{graphicx}
\usepackage{colortbl}
\usepackage{graphicx}
\usepackage{multirow}
\usepackage{amsmath}
\usepackage{placeins}
\usepackage{siunitx}
\usepackage{caption}
\usepackage{subcaption}
\usepackage{wrapfig}
\usepackage{afterpage}
\usepackage{adjustbox}
\usepackage{enumitem}

\title{\koala KoalaGS: Complexity-Aware Continuous\\Level of Detail for Gaussian Splatting}

\newcommand{\koala}{%
  \leavevmode\llap{\smash{\raisebox{-22.3pt}{\includegraphics[height=0.5in]{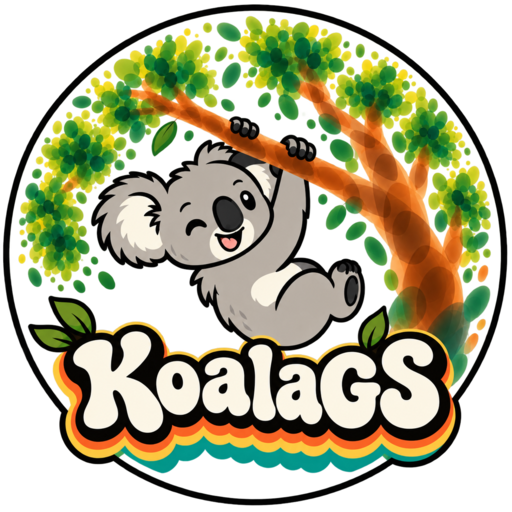}}}\hspace{0.35em}}%
}

\newcommand{\ie}{\textit{i.e.}}
\newcommand{\eg}{\textit{i.e.}}
\newcommand{\cf}{\textit{cf.}}

\newif\ifshowedits
\showeditsfalse

%% file: sections/00_introduction.tex
\vspace{-0.5em}
\section{Introduction}
\vspace{-0.5em}
\label{sec:intro}

3D Gaussian Splatting~\cite{kerbl20233dgs} (3DGS) has made neural scene representations practical for novel view synthesis in real time. As the representation has matured, deployment under heterogeneous bandwidth and compute budgets has motivated a growing line of work on \emph{continuous level of detail} (LoD) for 3DGS, where a single Gaussian set must remain useful when only a leading prefix of its primitives is retained~\cite{guo2026matryoshka, cheng2026clodgs, milef2025clod3dgs}. The central problem in this setting is determining which primitives
should retain capacity as the budget shrinks. Existing methods answer it implicitly, ranking primitives by opacity~\cite{guo2026matryoshka} or by learned importance scores~\cite{liu2025flexgs}. A structural property of the representation suggests a complementary signal. A Gaussian does not contribute equally throughout a scene, as a single primitive may account for a broad region of nearly uniform appearance yet resolve only a fraction of a visually intricate one. We contend that this spatial variation in the value of capacity is what a continuous LoD policy should exploit, and we show that it can be anticipated from the input views alone.

We investigate this claim through a pilot study. Prior work already showed that global image complexity predicts reconstruction error in neural scene representations~\cite{beche2025claravid}; however, continuous LoD requires a localized signal that identifies not only which views are difficult, but where that difficulty lies within the scene. We therefore measure the correlation between local image complexity and reconstruction quality in the 2D image plane and, after multiview projection, on the Gaussian primitives. The correlation persists across DL3DV~\cite{ling2024dl3dv} and multiple complexity functionals in both domains, supporting complexity as a spatial prior for reconstruction difficulty that is available before optimization. We then examine how 3DGS-MCMC allocates capacity with respect to this structure. Gaussian density increases with complexity, approximately exponentially in regions of high complexity, yet even visually simple regions are represented by many low-opacity Gaussians. Under truncation according to opacity, quality degrades at comparable rates in simple and complex regions, showing that opacity alone does not expose the spatial redundancy required for robust continuous LoD. Together, these observations expose a mismatch between how 3DGS-MCMC allocates capacity at full budget and how effectively that allocation is preserved under truncation.

Our continuous LoD method for Gaussian Splatting addresses this mismatch. The key insight is that image complexity, which is predictive of reconstruction error,
provides an inexpensive spatial prior for allocating capacity and selecting primitives as the budget shrinks. During optimization, a relocation rule biased by complexity steers primitives toward difficult regions, while at inference an importance score modulated by complexity preserves broad primitives in simple regions and relies primarily on opacity in complex ones. A residual variance objective further discourages an abrupt degradation as the budget shrinks. We evaluate the resulting representation along three complementary axes. Our method matches the best
baselines at full capacity,
while outperforming all continuous LoD baselines across the budget spectrum, even in extremely low budget ($<1\%$).
We show that this has great practical implication in streaming context,
where our method improves drastically the time-to-quality performance.

We summarize our contributions as follows:
\begin{itemize}[leftmargin=1.7em]
    \vspace{-0.3em}
    \item A pilot study on 140 DL3DV scenes showing image complexity predicts where reconstruction error concentrates, while capacity is allocated in a form that truncation cannot preserve.
    \vspace{-0.3em}
    \item A continuous LoD method combining relocation biased by complexity during training, importance scoring modulated by complexity at inference, and a residual variance objective that discourages localized degradation across budgets.
    \vspace{-0.3em}
    \item An evaluation of fixed budget quality, integrated spectrum quality, and transition variance, together with a streaming study showing a better time-to-quality tradeoff.
\end{itemize}

\vspace{-1.25em}

%% file: sections/01_related_work.tex
\section{Related Work}
\vspace{-1em}
\label{sec:rel_work}

\textbf{Gaussian Splatting.} 3D Gaussian Splatting  transformed neural rendering by using an explicit set of anisotropic Gaussian primitives optimized for visibility-aware rasterization, enabling high-fidelity novel view synthesis at real-time frame rates \cite{kerbl20233dgs}.
A large body of literature focuses on making this representation more stable, efficient, and deployable. On the optimization side, \cite{kheradmand20243dgsmcmc} reformulates densification as probabilistic sampling over primitives, \cite{lu2024scaffoldgs} introduces structured view-adaptive anchors, \cite{niemeyer2025radsplat} incorporates radiance-field-informed optimization for robust high-speed rendering, while \cite{mallick2024taming,fang2024minisplat,kotovenko2025edgs} reduce training cost or remove expensive densification mechanisms under constrained budgets. A complementary \cite{papantonakis2024reduce3dgs,fan2024lightgaussian} line compresses the trained representation itself, reducing memory footprint and rendering cost to make Gaussian scenes practical on resource-constrained devices. In parallel, other works \cite{huang20242dgs, held2025trianglegs, held2025trianglegsp} revisit the primitive parameterization, replacing or constraining 3DGS with 2D variants to improve geometric fidelity and compatibility with classical graphics pipelines. The flexibility of this representation also enables applications beyond static novel view synthesis, including dynamic scene modeling \cite{luiten2023dynamic,wu20244dgs}, progressive and adaptive streaming \cite{shi2025lapisgs,sun2025dynamiclapisgs}, mesh reconstruction \cite{guedon2024sugar}, or higher-level scene understanding through language grounding, segmentation, and editing \cite{qin2024langsplat,ye2024gaussiangroupping}. Altogether, these works establish 3DGS as a general explicit scene representation. Our work employs 3DGS for deployment under
heterogeneous primitive budgets, exposing image complexity as a spatial prior over both densification and ranking.

\textbf{Level of Detail in Gaussian Splatting.} A growing body of work ports classical Level of Detail \cite{hoppe1996progressivemeshes} into the splatting pipeline. A first category imposes \emph{explicit} hierarchies over the primitive set and exposes a small number of discrete operating points~\cite{ren2025octreegs, kerbl2024hierarchical, seo2025flod}, with spatial divide and conquer variants scaling to kilometer sized captures~\cite{liu2024citygaussian, lin2024vastgaussian, kulhanek2025lodge}. A second category approximates multi budget behavior through compression and importance pruning at varying targets~\cite{fang2024minisplat,fan2024lightgaussian}, where each operating point is optimized independently and a new model must be retrained for every budget. A third category pursues \emph{continuous} LoD over a single primitive set: \cite{cheng2026clodgs, milef2025clod3dgs} learn per Gaussian distance dependent opacity decay, while FlexGS~\cite{liu2025flexgs} learns a differentiable budget conditioned selector that achieves strong quality at targeted budgets but requires a fresh forward pass at every LoD change, precluding progressive streaming. Closer to our setting, Matryoshka-GS~\cite{guo2026matryoshka} introduces a nested prefix renderable ordering trained via stochastic budget sampling, so that any leading slice yields a coherent reconstruction. Our work belongs to this last category but departs from prior approaches in how the ordering is determined: we introduce image complexity as a scene side spatial prior over both densification and ranking, and train for spatial uniformity of the residual between full capacity and prefix renders.

%% file: sections/02_analysys_and_method.tex
\section{Pilot Study}
\label{sec:pilot}

\begin{figure}
    \centering
    \begin{subfigure}{0.9\linewidth}
        \includegraphics[width=\linewidth]{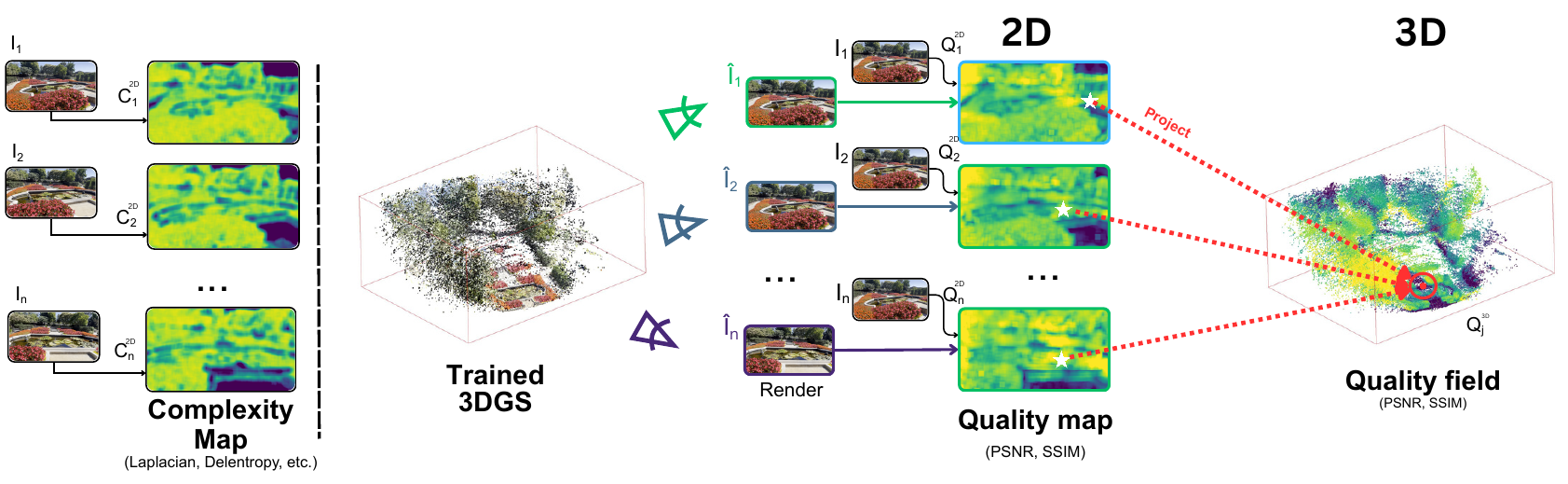}\vspace{-0.6em}
        \caption{Analysis methodology}
        \label{fig:complexity_meth}
    \end{subfigure}\\[0.3em]
    \begin{subfigure}{0.9\linewidth}
        \includegraphics[width=\linewidth]{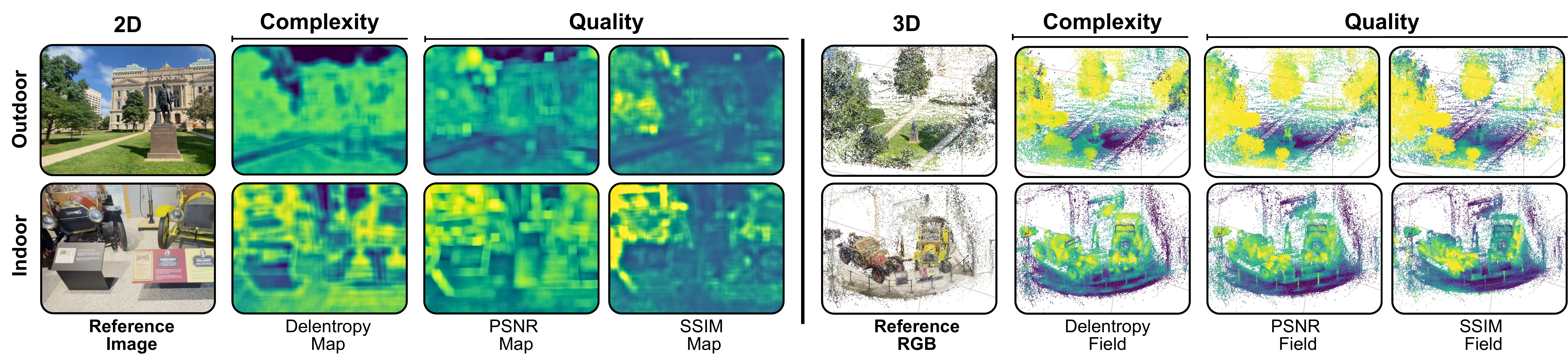}\vspace{-0.3em}

        \caption{Qualitative examples}
        \label{fig:complexity_examples}
    \end{subfigure}
    \caption{\textbf{Analysis methodology.} (\subref{fig:complexity_meth}) We measure the spatial correlation between the complexity map of the input image (left), and the quality map (center), obtained after 3DGS-MCMC optimization. We then project and aggregate these 2D fields onto the Gaussian particles to obtain complexity/quality fields in 3D (right). (\subref{fig:complexity_examples}) Complexity and quality inversely correlate both in the 2D image plane and in the 3D object space. \textit{Note that for easier visualization, we inverted colors of all complexity maps}.
\label{fig:complexityqual}}

    \vspace{-1em}
\end{figure}

This section studies image complexity as a predictor of 3DGS reconstruction quality, computed from the input views before any optimization. The outcomes will shape our method, which must decide where to spend capacity at the moment when reconstruction error, the signal that would otherwise guide it, is not yet available. The intuition is that deploying under a primitive budget is a spatial decision. As primitives are withdrawn, some regions tolerate the loss while others do not, so the capacity that remains should sit where reconstruction error concentrates. Image complexity addresses this, since prior work~\cite{beche2025claravid} shows that it predicts reconstruction error and is read from the input images before optimization rather than recovered after training. What that work leaves open is whether the signal is spatial, since an image level score ranks views by difficulty but not where the error lies within them. We therefore investigate two questions: \textbf{(i)} Does complexity localize reconstruction error in space, across the image plane and the 3D world? \textbf{(ii)} Are trained models already allocating capacity where complexity predicts it should? For experimental details we refer to Sec.~\ref{sec:supp_implementation:pilot_study}.

\paragraph{Does complexity spatially correlate with reconstruction error?} To address this, we first analyze the spatial distribution of complexity with respect to the reconstruction quality metrics, seeking to study how the two correlate both in the 2D image space and 3D space. The overall methodology is shown in~Fig.~\ref{fig:complexity_meth}. Specifically, we define two spatially indexed fields over image patches. Let $\hat I_i = \mathcal{R}_\theta(\Pi_i)$ denotes the rendering of a trained 3DGS-MCMC model under camera $\Pi_i$, and let $R(u) \subset \Omega$ be a receptive field centered at pixel $u$. For 2D, we define the quality map and complexity map as:
\begin{equation}
  Q^{\text{2D}}_i(u) = \frac{1}{|R(u)|}\sum_{v \in R(u)} q\!\left(I_i(v), \hat I_i(v)\right),
  \quad
  C^{\text{2D}}_i(u) = H\!\left(I_i\big|_{R(u)}\right),
  \label{eq:fields2d}
\end{equation}
where $H$ is a local complexity functional and $q$ is a local quality measure, with higher values indicating better reconstruction. To obtain their 3D counterparts, we project all quality maps $Q^{\text{2D}}_i$ and complexity maps $C^{\text{2D}}_i$ onto the Gaussian particles under multiview aggregation, obtaining a single quality field $Q^\text{3D}$ and complexity field $C^\text{3D}$ per scene. To do so, we define the contributing view set as $\mathcal{I}(\mathbf{x}) = \{\, i : \pi_i(\mathbf{x}) \in \Omega \ \text{and}\ \mathbf{x}\ \text{visible in view}\ i \,\}$, the views in which $\mathbf{x}$ projects inside the image domain and is not occluded, so that each point aggregates evidence only from views that observe it. Considering all 3D Gaussian positions $\mathbf{x}$, we then define the 3D fields via direct projection such that:
\begin{equation}
  Q^\text{3D}(\mathbf{x}) = \frac{1}{|\mathcal{I}|}\sum_{i \in \mathcal{I}}
    Q^{\text{2D}}_i\!\left(\pi_i(\mathbf{x})\right)
  \quad\text{and}\quad
  C^\text{3D}(\mathbf{x}) = \frac{1}{|\mathcal{I}|}\sum_{i \in \mathcal{I}}
    C^{\text{2D}}_i\!\left(\pi_i(\mathbf{x})\right).
  \label{eq:fields3d}
\end{equation}
In~Fig.~\ref{fig:complexity_examples} we report qualitative visuals on bounded and unbounded scenes. They highlight that regardless of the scene type, the complexity and quality fields generally correlate visually. This is very much noticeable on the outdoor pathway which are low complexity regions that also exhibit a low reconstruction error (high quality score).

We further measure quantitatively the statistical correlations of quality and complexity using the Pearson correlation $\rho$ which denotes the linear association, and the Spearman rank correlation $r_s$. For 2D, correlation is observed on paired observations $\{(C^{\text{2D}}_i(u),\, Q^{\text{2D}}_i(u))\}$ over all patches and views, while for 3D metrics we evaluate $\{C^\text{3D}(\mathbf{x}),\,Q^\text{3D}(\mathbf{x})\}$ over all Gaussian locations. Regardless of the measure used, results in Tab.~\ref{tab:correlation} show consistently high correlation, with large negative values indicating that highly complex regions consistently lead to low reconstruction quality. \emph{In the 2D image plane}, all three measures exhibit a consistent negative association with patch level quality across scenes and views. The effect is stronger under PSNR ($|\rho|, |r_s|$ in the range $0.55$--$0.65$) than SSIM ($0.21$--$0.50$). Among the functionals, \textit{delentropy} exhibits the strongest correlation, and we adopt it in our method. \emph{In the 3D object space}, the picture refines rather than uniformly attenuates. Compared to the 2D metrics, aggregation reduces view dependent appearance noise and recovers a clearer linear relationship (\ie, stronger Pearson correlation), whereas the rank correlations lower substantially ($|r_s| \approx 0.17$--$0.39$). The interpretation is that multi view averaging preserves the gross magnitude relationship between complexity and quality but compresses the tails of both fields, degrading the fine ordering even where the linear association strengthens. This empirical evidence supports that image complexity provides a spatial cue about \emph{where} reconstruction errors emerge, both within a view and within a scene. To further establish complexity as a usable spatial prior over reconstruction difficulty, we study how Gaussians are allocated with respect to complexity.

\begin{table}[t]
    \centering
    \scriptsize
    \setlength{\tabcolsep}{3.5pt}
    \renewcommand{\arraystretch}{0.92}
    \resizebox{0.6\linewidth}{!}{%
    \begin{tabular}{llccccc}
    \toprule
    & & \multicolumn{2}{c}{\textbf{Pearson $\rho$}}
      & \multicolumn{2}{c}{\textbf{Spearman $r_s$}} & \\
    \cmidrule(lr){3-4} \cmidrule(lr){5-6}
    \textbf{Domain} & \textbf{Complexity}
    & \textbf{PSNR} & \textbf{SSIM}
    & \textbf{PSNR} & \textbf{SSIM}
    & \textbf{Avg. Rank} \\
    \midrule
    \multirow{4}{*}{2D}
      & Shannon    & $-0.605$ & $-0.259$ & $-0.590$ & $-0.346$ & $3$ \\
      & Laplacian  & $-0.546$ & $-0.207$ & $\mathbf{-0.649}$ & $-0.365$ & $2$ \\
      & Delentropy & $\mathbf{-0.647}$ & $\mathbf{-0.379}$ & $-0.638$ & $\mathbf{-0.498}$ & $\mathbf{1}$ \\
      \cmidrule(l){2-7}
      & Average    & $-0.599$ & $-0.282$ & $-0.626$ & $-0.403$ & -- \\
    \midrule
    \multirow{4}{*}{3D}
      & Shannon    & $-0.588$ & $-0.557$ & $-0.170$ & $-0.165$ & $3$ \\
      & Laplacian  & $-0.655$ & $\mathbf{-0.688}$ & $-0.224$ & $-0.267$ & $2$ \\
      & Delentropy & $\mathbf{-0.678}$ & $-0.646$ & $\mathbf{-0.384}$ & $\mathbf{-0.388}$ & $\mathbf{1}$ \\
      \cmidrule(l){2-7}
      & Average    & $-0.640$ & $-0.630$ & $-0.259$ & $-0.273$ & -- \\
    \bottomrule
    \end{tabular}%
    }
    \vspace{0.3em}
    \caption{\textbf{Correlation between local complexity and reconstruction quality.}
    On DL3DV, across 140 scenes, complexity is negatively correlated with PSNR and SSIM in both the image plane and object space, indicating that more complex regions tend to reconstruct worse. Delentropy gives the strongest average rank. Best correlations are shown in bold.}
    \label{tab:correlation}
    \vspace{-1.75em}
\end{table}

\begin{figure}
    \centering
    \begin{subfigure}{0.51\linewidth}
        \includegraphics[width=1\linewidth]{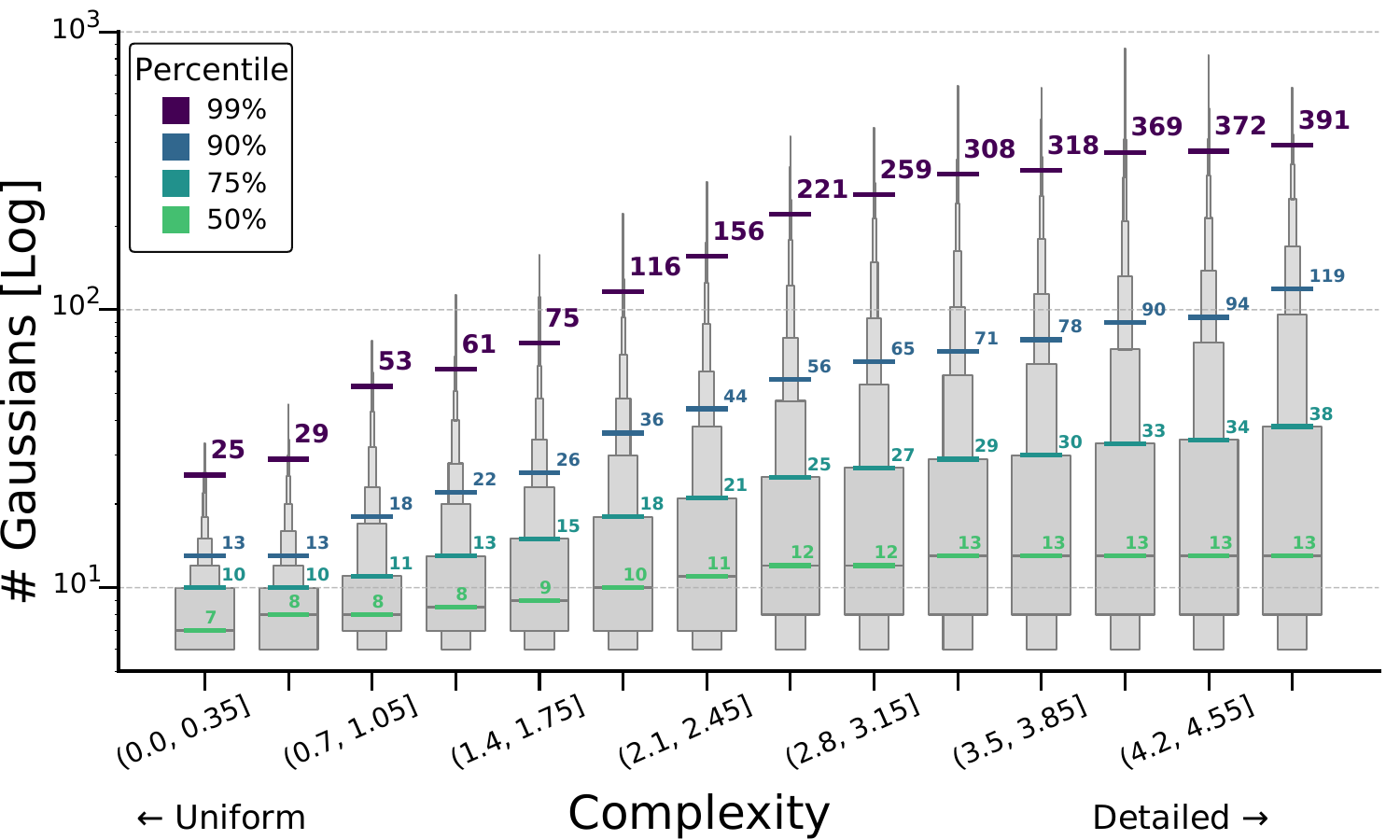}
        \caption{Number of Gaussians (log-scale) per complexity}
        \label{fig:complexity_allocation}
    \end{subfigure}
    \begin{subfigure}{0.48\linewidth}
        \includegraphics[width=1\linewidth,trim=0 0 0 0cm,clip]{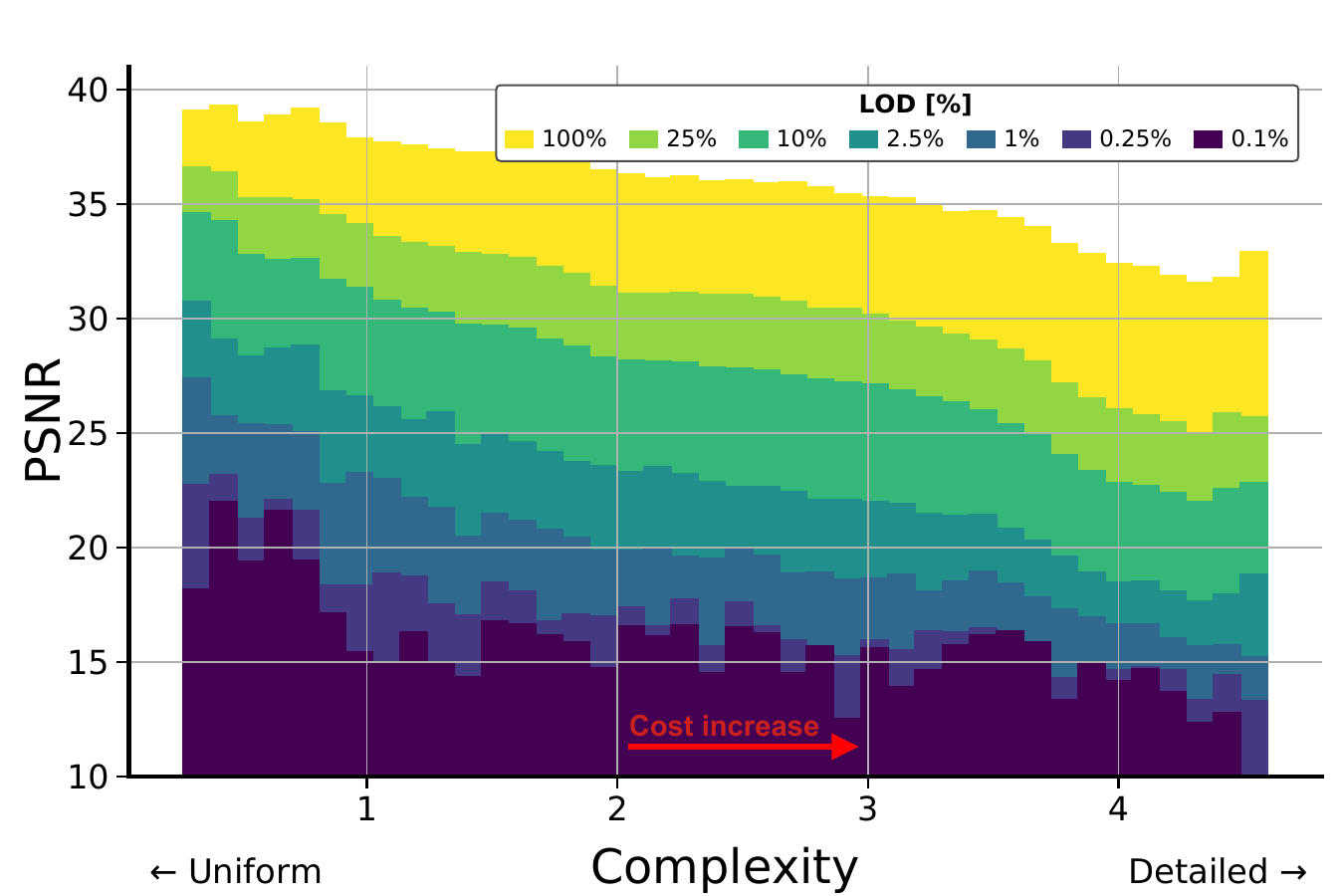}
        \caption{Quality (PSNR $\uparrow$) for varying LOD}
        \label{fig:complexity_lod}
    \end{subfigure}
    \vspace{-0.2em}
    \caption{\textbf{Relation between complexity and Gaussian density.} (\subref{fig:complexity_allocation}) Each bin counts the Gaussians inside a voxel of that complexity aggregated across DL3DV scenes. On the y-axis, the straight rise of the percentiles indicates an exponential growth (note y-axis is log-scaled). (\subref{fig:complexity_lod}) Considering an indoor scene reconstructed with 3DGS-MCMC, we lower capacity by dropping primitives in order of opacity, plotting the average PSNR reached at each complexity for every budget. We observe that the simplest regions lose almost as much quality as the most complex ones, so even flat surfaces are broken into multiple low opacity Gaussians.}
    \label{fig:complexitygaussians}
    \vspace{-1.5em}
\end{figure}

\paragraph{Does complexity correlate with Gaussian density?} Since complexity correlates with the reconstruction error, it is reasonable to expect that the optimal `per region number of Gaussians' is already achieved. Indeed, density control mechanism (pruning, cloning, relocation) for 3DGS are designed to increase Gaussian density where reconstruction error is higher.
To relate density to complexity at the scene level, we voxelize each scene at a fixed edge length of $0.1$ in normalized scene coordinates and, within every voxel, record the mean complexity, the mean reconstruction quality, and the number of Gaussians whose centers fall inside. The fixed scale is well defined because DL3DV test scenes are released in a comparable spatial scale across scenes.

Reporting the resulting Gaussian counts across complexity intervals in~Fig.~\ref{fig:complexity_allocation}, we observe that more complex voxels hold more Gaussians, generally following an exponential trend. At the 90th percentile, the ratio between the least and most complex voxels (13 vs. 119 Gaussians per voxel) is surprising, since the simplest voxels typically depict uniform color and are expected to be reconstructed with only a few Gaussians. We therefore question whether the Gaussian budget is correctly utilized across regions of varying complexity.

To verify this, we implement a naive LoD strategy which preserves only the top X\% of Gaussians by opacity. Varying LoD from 100\% (\ie, full capacity) to 0.1\%, in~Fig.~\ref{fig:complexity_lod} we report the PSNR across complexity for a representative scene. While overall PSNR decreases expectedly with complexity and LoD, the study reveals another intriguing characteristic. Decreasing LoD from $100\%$ to $0.1\%$ lowers quality nearly equally in complex regions ($\approx-62\%$ PSNR) and in simple regions ($\approx-58\%$ PSNR). This indicates that, even in the simplest regions, 3DGS-MCMC favors decomposing visual signals into multiple \textit{lower-opacity} Gaussians. Moreover, PSNR fluctuations across complexity reveal uneven Gaussian opacities. Altogether, these observations show that while 3DGS-MCMC reconstructions are denser in complex regions, Gaussians remain suboptimally distributed across complexities. We further analyze how Gaussians are allocated and retained across complexity levels in Sec.~\ref{supp:subsec_degradation_across_complexity_levels}.

\section{Method}
\vspace{-1em}

We consider the continuous level of detail setting for 3DGS.
Given a posed training set $\mathcal{D}=\{(I_i,\Pi_i)\}_{i=1}^{M}$, the goal is to learn a single Gaussian representation  $\mathcal{G}=\{g_j\}_{j=1}^{N}$ that can be rendered under a variable primitive budget $\tau\in(0,1]$, such that

$\mathcal{G}_{\tau}\subseteq\mathcal{G}$ with $|\mathcal{G}_{\tau}|=\lceil \tau N\rceil$.
At every budgeted representation, our aim is that $\mathcal{G}_{\tau}$ remains spatially coherent and degrades gracefully as $\tau$ decreases.
Following our pilot study (Sec.~\ref{sec:pilot}) which demonstrates that image complexity is a surrogate of the reconstruction error, our aim is to benefit from the image complexity signal, readily accessible at training time. Specifically, we apply two simple modifications to the optimization process of state-of-the-art 3DGS-MCMC~\cite{kheradmand20243dgsmcmc}. First, we utilize complexity to foster (re)allocation of the Gaussians in regions with higher complexity (Sec.~\ref{sec:method:adc}).
Second, we introduce a content-aware importance score that retains simple regions at low budgets (Sec.~\ref{sec:method:score}). We complement them with a residual variance objective that encourages stable reconstruction error across space and LoDs.
Experiments show that our simple modification boost performance and efficiency. The proposed LoD is global and view independent, rather than conditioned on camera distance or screen resolution. The implications are further discussed in Sec.~\ref{sec:supp:lod}.

\paragraph{Preliminaries.}
A 3DGS representation encodes a scene as a set of Gaussian primitives, where each Gaussian $g_j=(\mu_j,\alpha_j,\Sigma_j,\mathbf{c}_j)$ is parameterized by its position $\mu_j$, opacity $\alpha_j$, covariance $\Sigma_j$, and color coefficients $\mathbf{c}_j$. To expose complexity to the optimization process, we augment each primitive with a scalar variable $C_j$, which stores the local image complexity associated with the current Gaussian. This relates to the complexity field in Eq.~\ref{eq:fields3d}, but differs from the post-hoc analysis in our pilot study since, during training, Gaussian positions and supports change, and therefore their complexity values must be updated online.

To address this, after each full-capacity training render of a view $\Pi_i$, we project the means of the contributing Gaussians into the image plane, and update each particle's complexity via an exponential
moving average, using the complexity map $C^{\text{2D}}_i(u)$ from Eq.~\ref{eq:fields2d}, computed with delentropy. So that:
\begin{equation}
  C_j \;\leftarrow\; (1 - \eta)\,C_j \;+\; \eta\,C^{\text{2D}}_i\!\bigl(\pi_i(\mu_j)\bigr),
  \qquad \eta \in (0, 1].
  \label{eq:complexity_update}
\end{equation}
Only particles whose centers project in the current view's frustum are updated. Over the course of training each particle's $C_j$ converges to the multi-view average similar to those in Eq.~\eqref{eq:fields3d}.
At last, $S_j$ is the importance score of a particle used for prefix sorting computed using the scoring function $S(\cdot)$.

\subsection{Complexity-aware Adaptive density control}
\label{sec:method:adc}

In practice, 3DGS variants rely on opacity as a proxy for this error signal: during optimization new primitives are concentrated where opacity has accumulated, and regions the optimizer has not yet resolved remain under-served precisely because they have not been resolved. In 3DGS-MCMC~\cite{kheradmand20243dgsmcmc}, the probability $p$ to relocate at the coordinate of $\mu_j$ writes: $p(g_j) = \alpha_j$.
Given that complexity is a surrogate of reconstruction errors (\cf, Tab.~\ref{tab:correlation}), we simply bias this sampling weight towards more complex regions. We therefore sample relocation destinations with:
\begin{equation}
  p(g_j) = \alpha_j \cdot \exp\!\bigl(\tilde{C}_j\bigr)\,,
  \label{eq:adc_proposal}
\end{equation}
where $\tilde{C}_j\in [0,1]$ is the min-max normalized $C_j$.
Note that we utilize $\exp(\cdot)$ to mimic the exponential growing behavior of particles count with complexity, as in the 90th percentile in Fig.~\ref{fig:complexity_allocation}. Importantly, the multiplicative form keeps opacity in the loop since a primitive that has earned its keep should not be evicted for exploring other regions. To prevent collapse onto fixed complex regions, we apply \emph{complexity dropout}, setting $\tilde{C}_j=0$ with probability $\delta=0.2$ (ablated in Sec.~\ref{supp:subsec_drout_ablation}) and thus scoring that fraction of relocation candidates by opacity alone ($\delta{=}0$: none; $\delta{=}1$: all).

\subsection{Complexity-aware Level of Details}
\label{sec:method:score}

As mentioned in Sec.~\ref{sec:pilot}, under a limited budget (\ie, $\tau < 1$), only Gaussians with the highest importance score $S(\cdot)$ are preserved. In the literature, this typically resolves to using opacity~\cite{guo2026matryoshka}, \ie, $S(g_j) = \alpha_j$. As shown in Fig.~\ref{fig:complexity_lod}, this degrades low-complexity regions as much as complex ones, although they should be the cheapest to retain, because opacity captures learned contribution but not the total spatial footprint over which a primitive contributes. We therefore rank primitives by combining three complementary signals: opacity, scale, and complexity:
\begin{equation}
S(g_j) = \alpha_j + \lambda \cdot \frac{s(g_j)}{\exp\bigl(\tilde{C}_j(\mu_j)\bigr)},
\label{eq:score}
\end{equation}
where $s(g_j)$ is the max-normalized scale of $g_j$ and $\lambda = 2$ is a fixed scalar weight. The three components play distinct, complementary roles. Opacity captures the importance learned through the photometric objective, allowing a small high-opacity Gaussian to outrank a moderate-opacity one that covers a broader region. Scale supplies this missing spatial footprint, while complexity controls where it should influence the score. In high-complexity regions, $\exp(C)$ attenuates the scale term, avoiding preference for large, potentially blurring primitives and reducing the score toward opacity, which is most reliable where optimization pressure is concentrated. In low-complexity regions, where preserving broad spatial support matters more,
the second term reduces to $\lambda \cdot s$ as $\exp(C) \to 1$, restoring weight to large primitives that efficiently cover simple areas despite their moderate opacity.

We note that LoD-awareness requires the optimizer to vary LoD (\ie, $\tau$) during training. However, the scoring alone does not guarantee that the resulting render degrades uniformly since the optimization support varies with $\tau$.
Namely, at full-capacity ($\tau = 1$) the optimizer has access to all primitives to fit the image, while at lower LODs ($\tau < 1$) the optimizer only accesses the surviving primitives (according to~Eq.~\ref{eq:score}) and may push the latter away from their full-capacity solution.
To keep the resulting deviation spatially uniform, we consider the residual of the reconstruction error between full-capacity render and lower LODs render, and minimize it's variance.

Formally, at each step we compute the residual of the reconstruction error $\mathcal{E}(\cdot)$ at $\tau$ and at full-capacity: $\Delta_i = \mathcal{E}(I_i^{\text{full}}) - \mathcal{E}(I_i^{\tau})$, and minimize its spatial variance:
\begin{equation}
  \mathcal{L}_{\text{var}} \;=\; \mathrm{Var}_{u \in \Omega}\!\bigl(\Delta_i(u)\bigr).
  \label{eq:var}
\end{equation}

\paragraph{Training objectives.} We use the L1+D-SSIM loss of 3DGS-MCMC on the full ($\tau=1$) and budgeted ($\tau<1$) renders, add $\lambda_{\text{var}}\mathcal{L}_{\text{var}}$, and average them. Training details are given in Sec.~\ref{supp:training_details}.

%% file: sections/03_results.tex
\section{Experimental Results}
\vspace{-1em}
\label{sec:experimental_results}

We evaluate our method on three datasets~\cite{hedman2018deepblending,barron2022mipnerf,knapitsch2017tandt} and compare its performance against five baselines~\cite{guo2026matryoshka, liu2025flexgs,fan2024lightgaussian,ren2025octreegs,kheradmand20243dgsmcmc}.
Since performance of LoD-aware 3DGS may vary across LoD, it is not possible to characterize its performance by a single metric. We therefore assess our method along three complementary axes: \emph{quality} at predefined LoD targets (Sec.~\ref{subsec:absolute}), \emph{spectrum quality} as the per-metric area under the curve over the full LoD sweep (Sec.~\ref{subsec:decay}), and \emph{intra- and inter-LOD transition variance} (Sec.~\ref{subsec:eval_variance}), which together capture whether quality is high, whether it is sustained across the spectrum, and whether it degrades gracefully rather than abruptly.
We further conduct an ablation in Sec.~\ref{sec:ablation} to isolate the contribution of each component of our method. We refer the reader to Sec.~\ref{supp:experimental_evaluation} for the experimental setup description and Sec.~\ref{supp:sec_additional_exp} for additional experiments. In all tables, best results are in \textbf{bold} and second best are \underline{underlined}.

\subsection{Quality at Predefined LOD Targets}
\label{subsec:absolute}

\begin{table}[t]
\centering
\resizebox{0.99\textwidth}{!}{
\setlength{\tabcolsep}{3pt}
\newcommand{\best}[1]{\textbf{#1}}
\newcommand{\snd}[1]{\underline{#1}}
\begin{tabular}{ll c ccc ccc ccc ccc}
\toprule
& & & \multicolumn{3}{c}{\textbf{100\% Capacity}} & \multicolumn{3}{c}{\textbf{10\% Capacity}} & \multicolumn{3}{c}{\textbf{1\% Capacity}} & \multicolumn{3}{c}{\textbf{0.1\% Capacity}} \\
\cmidrule(lr){4-6} \cmidrule(lr){7-9} \cmidrule(lr){10-12} \cmidrule(lr){13-15}
& \textbf{Method} & \textbf{\#GS [k]} &
\begin{tabular}{@{}c@{}}\textbf{PSNR} $\uparrow$\end{tabular} &
\begin{tabular}{@{}c@{}}\textbf{SSIM} $\uparrow$\end{tabular} &
\begin{tabular}{@{}c@{}}\textbf{LPIPS} $\downarrow$\end{tabular} &
\begin{tabular}{@{}c@{}}\textbf{PSNR} $\uparrow$\end{tabular} &
\begin{tabular}{@{}c@{}}\textbf{SSIM} $\uparrow$\end{tabular} &
\begin{tabular}{@{}c@{}}\textbf{LPIPS} $\downarrow$\end{tabular} &
\begin{tabular}{@{}c@{}}\textbf{PSNR} $\uparrow$\end{tabular} &
\begin{tabular}{@{}c@{}}\textbf{SSIM} $\uparrow$\end{tabular} &
\begin{tabular}{@{}c@{}}\textbf{LPIPS} $\downarrow$\end{tabular} &
\begin{tabular}{@{}c@{}}\textbf{PSNR} $\uparrow$\end{tabular} &
\begin{tabular}{@{}c@{}}\textbf{SSIM} $\uparrow$\end{tabular} &
\begin{tabular}{@{}c@{}}\textbf{LPIPS} $\downarrow$\end{tabular} \\
\midrule
\multirow{6}{*}{\rotatebox{90}{\textbf{DB}}}
& OctreeGS \cite{ren2025octreegs}& 1,457 & 30.23 & 0.908 & 0.254 & - & - & - & - & - & - & - & - & - \\
& FlexGS \cite{liu2025flexgs}& 2,547 & 29.29 & 0.898 & 0.320 & 29.27 & 0.897 & 0.349 & 24.68 & 0.824 & 0.472 & - & - & - \\
\cmidrule{2-15}
& 3DGS MCMC \cite{kheradmand20243dgsmcmc}& 2,950 & 29.70 & \snd{0.907} & \best{0.236} & 24.36 & 0.792 & 0.362 & 16.81 & 0.656 & 0.520 & 12.67 & 0.601 & 0.599 \\
& LightGaussian \cite{fan2024lightgaussian} & 2,950 & \snd{29.73} & \snd{0.907} & \snd{0.240} & 26.21 & 0.849 & \snd{0.333} & 19.59 & 0.742 & 0.482 & 15.26 & 0.675 & 0.546 \\
& MGS \cite{guo2026matryoshka} & 2,950 & 29.29 & 0.905 & 0.247 & \snd{28.61} & \snd{0.888} & \best{0.282} & \snd{23.28} & \best{0.800} & \best{0.416} & \snd{16.58} & \snd{0.692} & \snd{0.531} \\
\rowcolor{gray!10}\cellcolor{white} & Ours & 2,950 & \best{29.92} & \best{0.911} & \snd{0.240} & \best{29.05} & \best{0.890} & \best{0.282} & \best{23.40} & \snd{0.798} & \snd{0.419} & \best{17.81} & \best{0.720} & \best{0.522} \\
\midrule
\multirow{6}{*}{\rotatebox{90}{\textbf{MipNerf}}}
& OctreeGS \cite{ren2025octreegs} & 2,991 & 29.16 & 0.876 & 0.174 & - & - & - & - & - & - & - & - & - \\
& FlexGS \cite{liu2025flexgs}& 3,012 & 29.06 & 0.872 & 0.217 & 27.58 & 0.835 & 0.293 & 23.39 & 0.663 & 0.494 & - & - & - \\
\cmidrule{2-15}
& 3DGS MCMC \cite{kheradmand20243dgsmcmc} & 3,092 & \snd{29.80} & \snd{0.893} & \best{0.146} & 20.58 & 0.647 & 0.352 & 14.14 & 0.383 & 0.586 & 11.06 & 0.267 & 0.667 \\
& LightGaussian \cite{fan2024lightgaussian} & 3,092 & \best{29.89} & \best{0.897} & \snd{0.147} & 21.06 & 0.669 & 0.359 & 15.81 & 0.475 & 0.558 & 13.01 & 0.377 & 0.636 \\
& MGS \cite{guo2026matryoshka}& 3,092 & 29.55 & 0.890 & 0.151 & \snd{26.97} & \snd{0.797} & \snd{0.280} & \snd{20.17} & \snd{0.544} & \snd{0.521} & \snd{14.49} & \snd{0.393} & \snd{0.620} \\
\rowcolor{gray!10}\cellcolor{white} & Ours & 3,092 & \snd{29.80} & \snd{0.893} & 0.150 & \best{27.82} & \best{0.822} & \best{0.258} & \best{21.14} & \best{0.566} & \best{0.510} & \best{16.04} & \best{0.459} & \best{0.611} \\
\midrule
\multirow{6}{*}{\rotatebox{90}{\textbf{TandT}}}
& OctreeGS \cite{ren2025octreegs} & 2,080 & 24.45 & 0.860 & 0.158 & - & - & - & - & - & - & - & - & - \\
& FlexGS \cite{liu2025flexgs} & 1,979 & 23.78 & 0.847 & 0.204 & 23.06 & 0.807 & 0.278 & 19.49 & 0.642 & 0.478 & - & - & - \\
\cmidrule{2-15}
& 3DGS MCMC \cite{kheradmand20243dgsmcmc} & 1,850 & 24.43 & \snd{0.866} & \best{0.143} & 18.52 & 0.670 & 0.356 & 11.96 & 0.404 & 0.589 & 8.92 & 0.306 & 0.680 \\
& LightGaussian \cite{fan2024lightgaussian} & 1,850 & \snd{24.46} & 0.863 & \snd{0.146} & 19.17 & 0.680 & 0.347 & 13.21 & 0.453 & 0.596 & 9.52 & 0.351 & 0.686 \\
& MGS \cite{guo2026matryoshka}& 1,850 & 24.40 & 0.864 & 0.150 & \snd{22.16} & \snd{0.755} & \snd{0.309} & \snd{16.66} & \snd{0.536} & \snd{0.526} & \best{12.10} & \snd{0.410} & \best{0.639} \\
\rowcolor{gray!10}\cellcolor{white} & Ours & 1,850 & \best{24.65} & \best{0.868} & 0.150 & \best{22.97} & \best{0.785} & \best{0.277} & \best{17.38} & \best{0.551} & \best{0.519} & \snd{11.56} & \best{0.413} & \snd{0.660} \\
\bottomrule
\end{tabular}
}
\vspace{0.3em}
\caption{\textbf{Performance at log-spaced LoD targets.} Metrics at 100\%, 10\%, 1\%, and 0.1\% budgets across three
datasets. Among full-LoD methods, ours performs on par at full capacity and generally better at lower budgets. For partial-LoD methods, dashes (-) denote unreachable operating points.}
\vspace{-1em}
\label{tab:lod_absolute}
\end{table}

\begin{figure}[t]
    \vspace{-0.5em}
    \centering
    \includegraphics[width=0.96\linewidth]{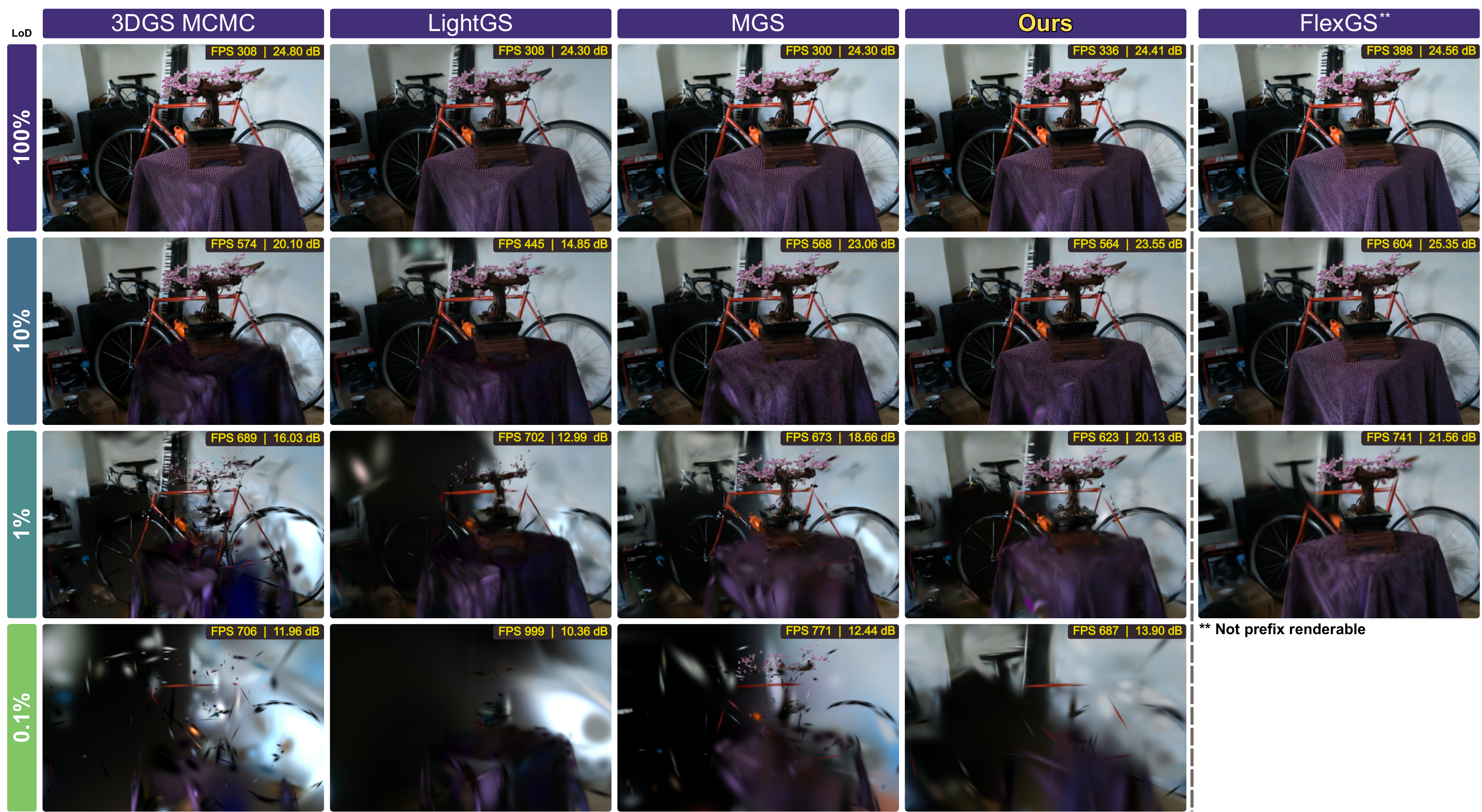}
        \caption{\textbf{Qualitative comparison across the LoD spectrum.} Methods perform similarly at full capacity but diverge sharply at lower budgets. Ours better preserves scene coverage at low budgets and remains competitive with FlexGS while enabling prefix rendering.}
    \label{fig:visual_quality}
    \vspace{-0.25em}
\end{figure}

\begin{figure}
    \vspace{-0.5em}
    \centering
    \includegraphics[width=0.86\linewidth]{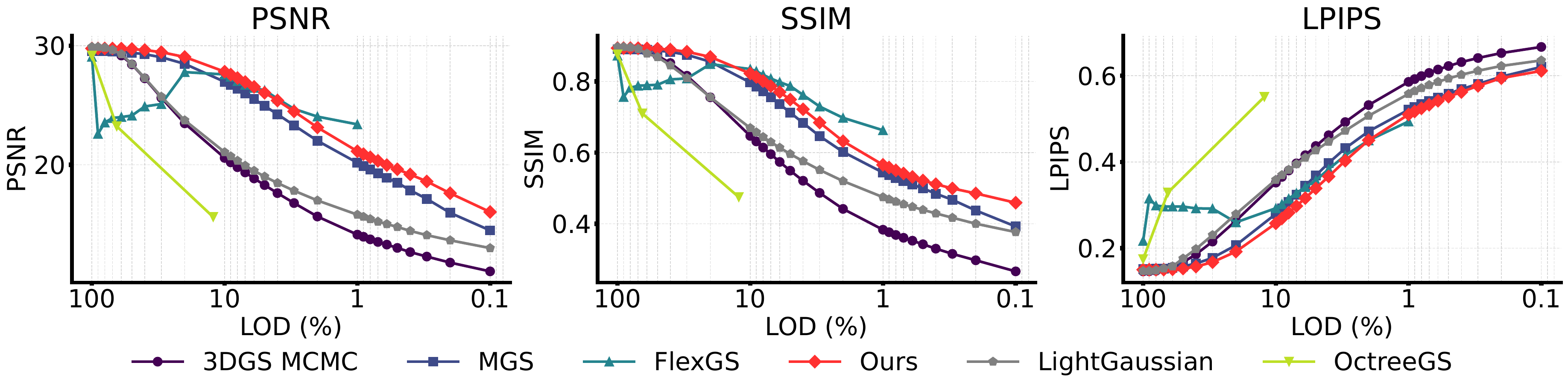}
    \caption{\textbf{Quality metrics across the LoD spectrum.} Our method matches full-capacity baselines and degrades more gracefully as budgets shrink. Below 10\%, 3DGS MCMC collapses and LightGaussian improves little. OctreeGS has limited discrete LoDs, FlexGS fluctuates between trained budgets, and MGS follows ours but with lower quality throughout. Results are on MipNerf.}
    \vspace{-1em}
    \label{fig:decay}
\end{figure}

We first ask whether our method delivers high quality at the operating points a user is most likely to deploy, therefore evaluating at four log-spaced budgets chosen to probe each decade of the spectrum: 100\% reflects full-capacity quality, 10\% a moderate compression regime relevant to deployment, 1\%~an aggressive reduction where ranking quality dominates and an extreme 0.1\%, [typically used when initialising a streaming]. Quantitative results are reported in Tab.~\ref{tab:lod_absolute}, while separating partial-LoD methods (OctreeGS, FlexGS) as they can only be evaluated at some intervals. Qualitatives are presented in Fig.~\ref{fig:visual_quality}, and curves are visualized in Fig.~\ref{fig:decay}. A consistent picture emerges across datasets. \emph{At full capacity}, all methods are competitive (Fig.~\ref{fig:visual_quality}, top row), with our method performing generally on par with the strongest baseline on every dataset and confirming that content-aware modifications introduce no penalty at peak fidelity.
\emph{As the budget shrinks}, the field separates sharply. 3DGS MCMC degrades rapidly and has already collapsed by the 1\% threshold, confirming that opacity-based ranking alone cannot sustain quality under reduction. {LightGaussian} retains more quality than opacity-based pruning in most cases, but falls behind both MGS and ours at every reduced budget across all three datasets. OctreeGS cannot operate below its native discrete levels and it remains competitive only on the indoor Deep Blending scenes and falls behind on outdoor scenes. FlexGS, which is not prefix renderable, trades this property for higher quality near the specific operating points it was optimised for, but degrades unpredictably between them. MGS, our closest competitor, follows the same trend as our method but consistently lags behind it. Altogether, content-aware ranking yields a more robust quality versus quantity trade-off than any single-family representative, both quantitatively and visually. We report the rendering speeds in Sec.~\ref{supp:subsec_rendering_speed} and further investigate how our importance score and densification behave jointly in Sec.~\ref{supp:subsec_degradation_across_complexity_levels}.

\vspace{-0.75em}
\subsection{Quality Across the LoD Spectrum}
\label{subsec:decay}
The previous section leaves open how quality evolves \textit{across the full LoD spectrum}. To address this, we sample 28 operating points in $[0.1\%, 100\%]$, ten linearly spaced per interval ($[0.1\%, 1\%]$, $[1\%, 10\%]$, $[10\%, 100\%]$) with shared endpoints, and integrate per-metric curves with the trapezoidal rule to compute interval-wise area-under-the-curve (AUC) score, rescaled per interval for direct comparison. Since discrete methods cannot operate across the full spectrum, we focus on prefix-renderable continuous LoD methods. From results in Tab.~\ref{tab:auc_intervals}, our method consistently achieves the highest AUC.
Since the total AUC is dominated by high budgets where all methods cluster, the sub-interval breakdown is more informative. {3DGS MCMC} collapses below 10\% capacity, yielding the lowest sub-interval AUC across datasets and confirming that opacity-based ranking alone fails as the budget shrinks. {LightGaussian} improves PSNR and generally SSIM over opacity-based pruning, but remains behind MGS and ours across all metrics in the two lower-budget intervals. Our largest gain over MGS occurs precisely in the low-budget $[0.1\%, 1\%]$ regime where ranking quality dominates. FlexGS performs better at $[1\%, 10\%]$, but it does not allow full LoD coverage.
Besides, FlexGS is outperformed by ours in other intervals, which directly results from its design as it trained for specific operating points at the cost of global consistency, producing the irregular degradation curves visible in Fig.~\ref{fig:decay}.
Overall, these results indicate that our method offers the most balanced quality versus quantity trade-off across the LoD spectrum.
\begin{table*}[t]
\centering
\resizebox{0.95\textwidth}{!}{
\setlength{\tabcolsep}{3pt}
\newcommand{\best}[1]{\textbf{#1}}
\newcommand{\snd}[1]{\underline{#1}}
\begin{tabular}{ll ccc ccc ccc ccc}
\toprule
& & \multicolumn{3}{c}{\textbf{Total [0.1\%, 100\%]}} & \multicolumn{3}{c}{\textbf{[0.1\%, 1\%]} ($\times 100$)} & \multicolumn{3}{c}{\textbf{[1\%, 10\%]} ($\times 10$)} & \multicolumn{3}{c}{\textbf{[10\%, 100\%]}} \\
\cmidrule(lr){3-5} \cmidrule(lr){6-8} \cmidrule(lr){9-11} \cmidrule(lr){12-14}
 & \textbf{Method} &
\begin{tabular}{@{}c@{}}\textbf{AUC} \\ \textbf{PSNR} $\uparrow$\end{tabular} &
\begin{tabular}{@{}c@{}}\textbf{AUC} \\ \textbf{SSIM} $\uparrow$\end{tabular} &
\begin{tabular}{@{}c@{}}\textbf{AUC} \\ \textbf{LPIPS} $\downarrow$\end{tabular} &
\begin{tabular}{@{}c@{}}\textbf{AUC} \\ \textbf{PSNR} $\uparrow$\end{tabular} &
\begin{tabular}{@{}c@{}}\textbf{AUC} \\ \textbf{SSIM} $\uparrow$\end{tabular} &
\begin{tabular}{@{}c@{}}\textbf{AUC} \\ \textbf{LPIPS} $\downarrow$\end{tabular} &
\begin{tabular}{@{}c@{}}\textbf{AUC} \\ \textbf{PSNR} $\uparrow$\end{tabular} &
\begin{tabular}{@{}c@{}}\textbf{AUC} \\ \textbf{SSIM} $\uparrow$\end{tabular} &
\begin{tabular}{@{}c@{}}\textbf{AUC} \\ \textbf{LPIPS} $\downarrow$\end{tabular} &
\begin{tabular}{@{}c@{}}\textbf{AUC} \\ \textbf{PSNR} $\uparrow$\end{tabular} &
\begin{tabular}{@{}c@{}}\textbf{AUC} \\ \textbf{SSIM} $\uparrow$\end{tabular} &
\begin{tabular}{@{}c@{}}\textbf{AUC} \\ \textbf{LPIPS} $\downarrow$\end{tabular} \\
\midrule
\multirow{5}{*}{\rotatebox{90}{\textbf{DB}}}
& FlexGS \cite{liu2025flexgs} & 27.47 & 0.867 & 0.363 & - & - & - & {25.28} & {0.788} & 0.350 & 24.94 & 0.788 & 0.328 \\
\cmidrule{2-14}
& 3DGS MCMC \cite{kheradmand20243dgsmcmc} & 27.96 & 0.870 & 0.274 & 13.69 & 0.573 & 0.497 & 19.49 & 0.664 & 0.378 & 25.87 & 0.798 & 0.231 \\
& LightGaussian \cite{fan2024lightgaussian} & 28.61 & 0.888 & 0.268 & 16.29 & 0.650 & 0.457 & 21.44 & \snd{0.730} & 0.347 & \snd{26.31} & 0.809 & 0.229 \\
& MGS \cite{guo2026matryoshka} & \snd{28.95} & \snd{0.897} & \snd{0.259} & \snd{18.81} & \snd{0.685} & \snd{0.414} & \snd{24.62} & \best{0.778} & \snd{0.289} & 26.30 & \snd{0.813} & \snd{0.226} \\
\rowcolor{gray!10}\cellcolor{white} & Ours & \best{29.51} & \best{0.902} & \best{0.254} & \best{19.19} & \best{0.690} & \best{0.413} & \best{24.86} & \best{0.778} & \best{0.288} & \best{26.83} & \best{0.817} & \best{0.221} \\
\midrule
\multirow{5}{*}{\rotatebox{90}{\textbf{MipNerf}}}
& FlexGS \cite{liu2025flexgs} & 24.77 & 0.792 & 0.293 & - & - & - & {23.36} & {0.698} & 0.328 & 22.43 & 0.722 & 0.260 \\
\cmidrule{2-14}
& 3DGS MCMC \cite{kheradmand20243dgsmcmc} & 26.58 & 0.814 & 0.213 & 11.69 & 0.307 & 0.559 & 16.39 & 0.493 & 0.395 & 24.83 & 0.762 & 0.168 \\
& LightGaussian \cite{fan2024lightgaussian} & 26.78 & 0.820 & 0.218 & 14.75 & 0.438 & 0.592 & 19.03 & 0.596 & 0.430 & 24.93 & 0.762 & 0.174 \\
& MGS \cite{guo2026matryoshka} & \snd{28.65} & \snd{0.858} & \snd{0.192} & \snd{16.45} & \snd{0.445} & \snd{0.503} & \snd{22.27} & \snd{0.636} & \snd{0.334} & \snd{26.25} & \snd{0.790} & \snd{0.153} \\
\rowcolor{gray!10}\cellcolor{white} & Ours & \best{29.09} & \best{0.867} & \best{0.184} & \best{17.55} & \best{0.471} & \best{0.496} & \best{23.20} & \best{0.665} & \best{0.311} & \best{26.59} & \best{0.795} & \best{0.147} \\
\midrule
\multirow{5}{*}{\rotatebox{90}{\textbf{TandT}}}
& FlexGS \cite{liu2025flexgs} & 21.89 & 0.788 & 0.262 & - & - & - & {19.82} & {0.679} & {0.308} & 19.91 & 0.720 & 0.231 \\
\cmidrule{2-14}
& 3DGS MCMC \cite{kheradmand20243dgsmcmc} & 22.50 & 0.801 & 0.212 & 9.74 & 0.331 & 0.561 & 14.42 & 0.510 & 0.406 & 20.96 & 0.747 & 0.166 \\
& LightGaussian \cite{fan2024lightgaussian} & 22.74 & 0.802 & 0.211 & 10.66 & 0.376 & 0.570 & 15.30 & 0.534 & 0.394 & 21.11 & 0.745 & 0.166 \\
& MGS \cite{guo2026matryoshka} & \snd{23.59} & \snd{0.827} & \snd{0.201} & \snd{13.67} & \snd{0.442} & \snd{0.519} & \snd{18.28} & \snd{0.609} & \snd{0.353} & \snd{21.62} & \snd{0.762} & \snd{0.160} \\
\rowcolor{gray!10}\cellcolor{white} & Ours & \best{24.02} & \best{0.839} & \best{0.189} & \best{13.83} & \best{0.447} & \best{0.517} & \best{19.05} & \best{0.637} & \best{0.327} & \best{21.98} & \best{0.771} & \best{0.151} \\
\bottomrule
\end{tabular}
}
\caption{\textbf{Area Under the Curve (AUC) across Capacity Intervals.} For all intervals, our method outperforms all other prefix-renderable continuous LoD methods, exhibiting large gaps in smaller intervals $[0.1\%, 1\%]$. Intervals are scaled for direct comparison.}
\label{tab:auc_intervals}
\end{table*}

\vspace{-0.5em}
\subsection{LoD Transition Variance}
\label{subsec:eval_variance}

\begin{figure}[t]
    \centering
    \begin{subfigure}[t]{0.495\linewidth}
        \centering
        \caption{Inter-LOD variance analysis.}
        \includegraphics[width=\linewidth]{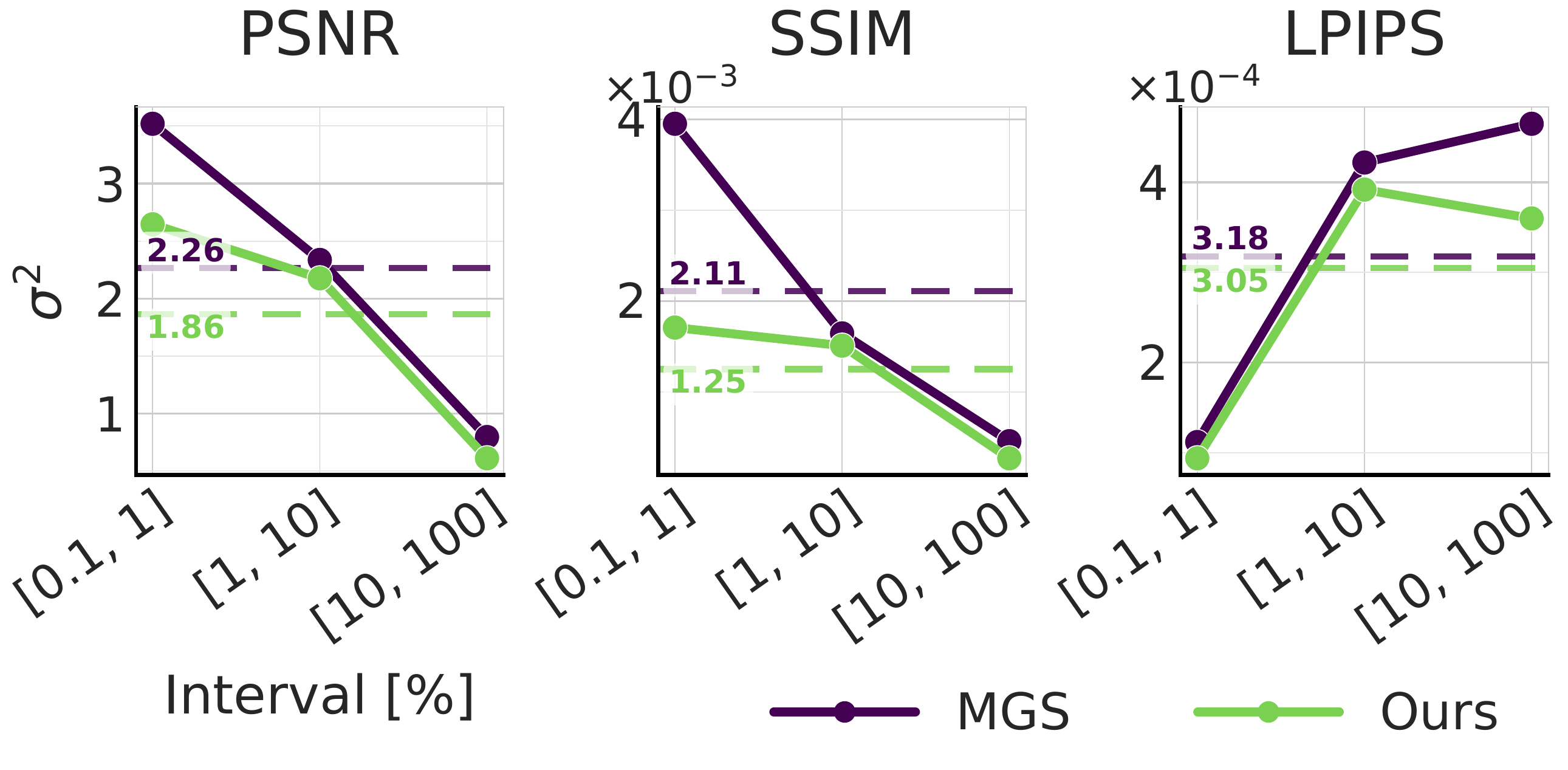}\vspace{-0.3em}
        \label{fig:inter_lod_var}
    \end{subfigure}\hfill
    \begin{subfigure}[t]{0.495\linewidth}
        \centering
        \caption{Intra-LoD variance analysis.}
        \includegraphics[width=\linewidth]{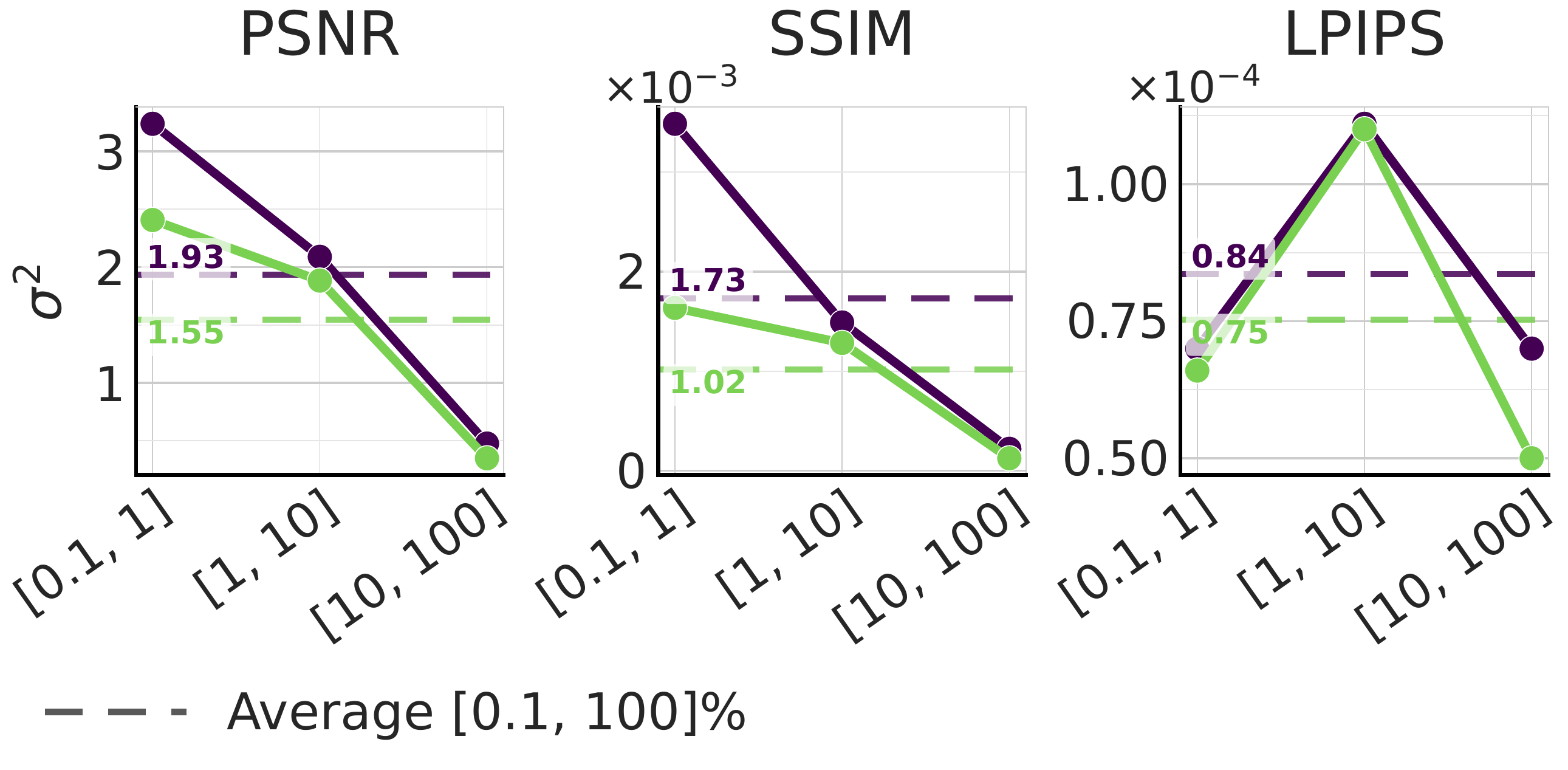}\vspace{-0.3em}
        \label{fig:intra_lod_var}
    \end{subfigure}
    \vspace{-0.8em}
    \caption{\textbf{Variance analysis across LoD.} For both MGS and ours, we report the inter and intra LoD variances on MipNerf(\textit{lower is better}). Our method consistently outperforms MGS.}
    \vspace{-1.5em}
    \label{fig:lod_variance}
\end{figure}

We further assess whether our method enables fine-grained reconstruction control with smooth quality degradation. Indeed, the above mentioned metrics do not capture spatial or LoD consistency: a method that concentrates errors in localized regions may rank highly while exhibiting abrupt changes across LoDs. To analyze this, we compute per-pixel error maps following Sec.~\ref{sec:pilot} and derive two complementary variance measures. \emph{Inter-LoD variance} captures transitions across levels: for each patch, we take differences between error maps at consecutive LoDs, compute the variance of these residuals, and average over all patches. Low values indicate smooth rather than abrupt degradation.
\emph{Intra-LoD variance} measures spatial uniformity within a level: we compute the variance of each error map across patches and average over LoDs. Low values indicate that errors are evenly distributed rather than concentrated in localized regions.
Neither measure is meaningful in isolation, as uniformly poor renderings can also yield low variance. We therefore use them as a \textit{complement} to the absolute quality metrics reported above. Further details are provided in Sec.~\ref{sec:lod_variance_measure}.
Results on MipNerf against MGS are reported in Fig.~\ref{fig:lod_variance}. Across all intervals and metrics, our method exhibits lower inter-LoD variance than MGS, with the gap being most pronounced in the $[0.1\%, 1\%]$ regime where budget pressure is highest and abrupt transitions are most likely to occur. A consistent reduction is also observed in intra-LoD variance, indicating that our content-aware ranking not only smooths transitions across levels but also distributes error more uniformly within each level. Taken together with the absolute and AUC results, these findings confirm that complexity-aware importance scoring yields a representation that degrades both more gracefully and more uniformly across the full LoD spectrum.

\begin{table}[t]
\centering
\resizebox{0.75\textwidth}{!}{
\begin{tabular}{l ccc ccc ccc}
\toprule
& \multicolumn{3}{c}{\textbf{Components}}
& \multicolumn{3}{c}{\textbf{AUC [0.1\%, 100\%]}}
& \multicolumn{3}{c}{\textbf{100\% Capacity}} \\
\cmidrule(lr){2-4} \cmidrule(lr){5-7} \cmidrule(lr){8-10}
& Dens. & Score & Var.
& PSNR $\uparrow$ & SSIM $\uparrow$ & LPIPS $\downarrow$
& PSNR $\uparrow$ & SSIM $\uparrow$ & LPIPS $\downarrow$ \\
\midrule
None & \xmark & \xmark & \xmark
& 28.74 & 0.855 & 0.196
& 29.55 & 0.863 & 0.164 \\
\midrule
\multirow{3}{*}{Single} & \cmark & \xmark & \xmark
& 28.81 & 0.859 & 0.190
& \underline{29.79} & \underline{0.889} & \underline{0.151} \\
& \xmark & \cmark & \xmark
& 28.94 & 0.861 & 0.188
& 29.62 & 0.882 & 0.159 \\
& \xmark & \xmark & \cmark
& 28.79 & 0.855 & 0.190
& 29.59 & 0.870 & 0.160 \\
\midrule
\multirow{3}{*}{Pair} & \cmark & \cmark & \xmark
& \underline{28.99} & \underline{0.862} & \underline{0.187}
& 29.75 & 0.885 & 0.154 \\
& \cmark & \xmark & \cmark
& 28.75 & 0.850 & 0.197
& 29.70 & 0.877 & 0.159 \\
& \xmark & \cmark & \cmark
& 28.95 & 0.860 & 0.189
& 29.71 & 0.880 & 0.155 \\
\midrule
\rowcolor{gray!10}
All & \cmark & \cmark & \cmark
& \textbf{29.09} & \textbf{0.867} & \textbf{0.184}
& \textbf{29.80} & \textbf{0.893} & \textbf{0.150} \\
\bottomrule
\end{tabular}
}\vspace{0.3em}
\caption{\textbf{Full ablation.} All combinations of complexity-aware densification (Dens.), importance scoring (Score), and the spatial variance loss (Var.) on MipNerf.}
\vspace{-1.0em}
\label{tab:ablation}
\end{table}

\subsection{Ablation Study}
\label{sec:ablation}
To isolate the individual and combined contributions of our three additions, we evaluate all combinations of content-aware densification (Dens.), content-aware importance scoring (Score), and the spatial variance loss (Var.). We report both full-capacity quality (100\%) and the per-metric AUC over the LoD spectrum $[0.1\%, 100\%]$ in Tab.~\ref{tab:ablation}. Each component alone improves upon the baseline across all AUC metrics, except Var., which matches the baseline in AUC SSIM. Densification provides the largest standalone improvement at full capacity, while importance scoring yields the largest standalone AUC gains. Among the two-component configurations, Dens.+Score performs best across both full-capacity and AUC metrics. However, adding Var.\ to either component alone does not consistently improve performance, suggesting that its benefit depends on the accompanying allocation and ranking mechanisms. The complete configuration achieves the best full-capacity quality and integrated quality across all three metrics, supporting the combined use of the three components. A more detailed breakdown of the ablation results at individual LoD budgets is provided in Sec.~\ref{sec:supp_ablation}.

\vspace{-0.5em}
\subsection{Application: time-to-quality in streaming}
\begin{figure}
    \centering
    \includegraphics[width=\linewidth]{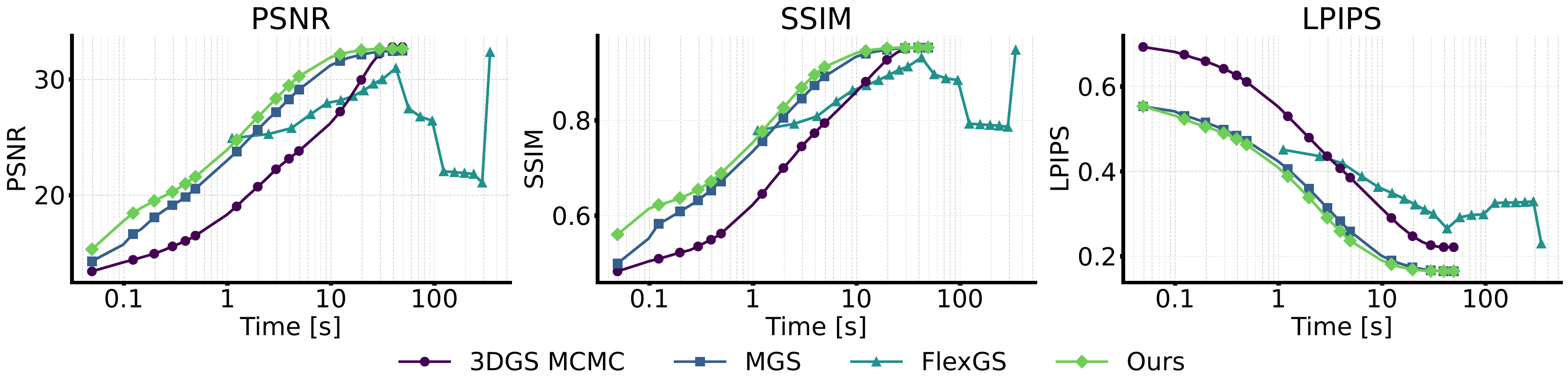}
    \vspace{-1.8em}
    \caption{\textbf{Time to quality in streaming.} Rendering quality of Bonsai~\cite{barron2022mipnerf}, over a $50$\,Mbps link. Prefix methods send each primitive once, so every arrival refines the model, and ours reaches every quality level first. FlexGS resends the full set for each budget, so time is delayed and quality oscillates.}
    \label{fig:ttq}
    \vspace{-1.1em}
\end{figure}

Our ranking naturally enables progressive streaming, where remote clients render the primitives as they arrive.
Prior 3DGS streaming systems use a few discrete quality layers~\cite{shi2025lapisgs} without characterizing how quality evolves throughout streaming.
We therefore measure \emph{time to quality}: the time to reach a target rendering fidelity at fixed bandwidth.
We simulate $50$\,Mbps bandwidth (representative of 5G network), and send each primitive as raw fp32 attributes ($236$ bytes).
We compare against MGS~\cite{guo2026matryoshka} and 3DGS MCMC~\cite{kheradmand20243dgsmcmc}, which stream incrementally like ours, and FlexGS~\cite{liu2025flexgs}, which must restream the full set at every budget.
On the Bonsai scene of MipNerf~\cite{barron2022mipnerf} (Fig.~\ref{fig:ttq}), the prefix methods receive the full model in $49.1$ s, while FlexGS needs $344$\,s, a $7.0\times$ overhead caused by restreaming. Our method reaches $25$\,dB after $1.47$\,s, compared to $1.96$\,s for MGS and $9.82$\,s for 3DGS MCMC, and $28$\,dB after $2.95$\,s, compared to $3.93$\,s and $14.73$\,s, respectively. FlexGS is also non monotone in time, dropping to about $22$\,dB in the mid budgets before recovering at full capacity. Because our importance score (Eq.~\ref{eq:score}) sends large primitives in simple regions first, the client sees a complete, coarse scene within the first timestamps.

%% file: sections/04_conclussion.tex
\vspace{-1.em}
\section{Conclusion}
\vspace{-1em}
\label{sec:conclusion}

We studied how to allocate Gaussian primitives as rendering budgets shrink in continuous LoD splatting. Our pilot study showed that image complexity, accessible before any optimization, predicts spatial reconstruction error in 2D and 3D, yet trained models fail to allocate capacity accordingly in a form that a shrinking budget preserves. Building on these observations, we design a method allocating primitives based on complexity which determines their survival at inference, using a variance objective to maintain uniform error across scales. Evaluated on three datasets, our approach matches the strongest baselines at full capacity, while achieving the highest integrated quality across budgets, and degrades more gracefully, with the largest gains below $1\%$ capacity and faster time-to-quality in streaming.

%% file: sections/05_supp.tex
\section{Limitations}

We identify three limitations that needs discussing of our approach, namely the optimization stability, the nature of our complexity prior, and view conditioning.\\
\textbf{(i)}~Our method inherits the stochastic relocation dynamics of 3DGS-MCMC, so results can slightly vary across runs. This variation does not alter our standing relative to prior work, as our method maintains its relative position at every capacity level and on every dataset we evaluate. The instability is moreover present to the same degree in 3DGS-MCMC and MGS, since all three build on the same relocation mechanism.\\
\textbf{(ii)}~The complexity prior we introduce is computed from the input images alone, and therefore captures spatial difficulty but neither view-dependent appearance nor semantic saliency. A specular highlight that shifts with viewpoint, or a small object that is salient despite low pixel complexity, receives no special treatment under our prior, and the correlations of Tab.~\ref{tab:correlation} should accordingly be read as holding under standard capture conditions.\\
\textbf{(iii)}~Primitive selection is likewise unconditioned on the viewpoint: the formulation is neither distance- nor resolution-dependent, and is thus not designed to suppress aliasing when rendering from distant viewpoints or at reduced resolution. However, its preference for large-support primitives in simple regions may reduce the retention of subpixel primitives relative to opacity-only ranking, but this provides no antialiasing guarantee. Because we modify primitive allocation and ordering without prescribing a competing filtering mechanism, our representation remains compatible in principle with dedicated approaches such as the smoothing filter of Mip-Splatting~\cite{yu2024mipsplatting}, a combination we do not evaluate here.

\section{Implementation details}
\label{sec:supp_implementation}

\subsection{Pilot Study}
\label{sec:supp_implementation:pilot_study}
\paragraph{Experimental setup.} We run the study on the DL3DV~\cite{ling2024dl3dv} test set of 140 scenes, reconstructed with 3DGS-MCMC~\cite{kheradmand20243dgsmcmc}, a robust 3DGS approach. Each scene is labeled bounded or unbounded from the DL3DV metadata and receives 1.0M primitives when bounded and 1.5M when unbounded, reflecting the larger extent of unbounded captures. All scenes are trained at 960p for 30k steps under the default 3DGS-MCMC hyperparameters, holding out every 8th frame as the test split. We measure reconstruction quality with PSNR~\cite{haskell1997psnr} and SSIM~\cite{wang2004ssim}, and image complexity with three functionals: Shannon pixel entropy~\cite{shannon1948mathematical,haskell1997psnr}, Laplacian response~\cite{haskell1997psnr}, and delentropy~\cite{larkin2016reflections}. Throughout our study, we exclude the particles that do not contribute to the final reconstruction.

\paragraph{3D aggregation of quality and complexity.} The 3D fields $Q^{3D}$ and $C^{3D}$ defined in Eq.~\ref{eq:fields3d} of the main paper are constructed by projecting the per-view 2D maps onto the trained Gaussian set under multi-view aggregation. To analyse the joint distribution of complexity, quality, and Gaussian density at the scene level (Fig.~\ref{fig:complexity_allocation}), we voxelize each scene using a fixed voxel size of $0.1$ in the Colmap normalised scene coordinates. This fixed scale is well defined because DL3DV test scenes are released in a roughly similar normalized frame, so a single voxel size yields comparable spatial granularity across scenes without per-scene tuning. Within each voxel we record (i) the mean local complexity, (ii) the mean reconstruction quality (PSNR or SSIM), and (iii) the number of Gaussians whose means fall inside.

\subsection{Method}
\label{supp:method_implementation}

\label{sec:supp:lod}
\paragraph{Scope of the LoD Formulation.}
Our formulation provides global, view-independent LoD through a single primitive ordering whose prefixes define successive approximations. Unlike camera-dependent selection over a spatial hierarchy, this ordering is fixed across viewpoints and budgets, enabling progressive streaming by transmitting additional primitives as the budget increases.

\textbf{Complexity maps computation.} The complexity maps are computed once per scene, before training, and cached to disk. The computation is implemented in PyTorch and runs on the GPU. Its cost depends on the complexity functional, the patch size, and the image resolution, and ranges from $20$\,s to $60$\,s per scene on the evaluated datasets. As this is a one time preprocessing step, its overhead is negligible relative to training. We use a fixed receptive field of $32$ pixels in all experiments, with a stride of $4$ to $8$ depending on the image resolution.  This setting yielded consistent complexity statistics across input resolutions ranging from 960p (DL3DV) to approx. 1.6K (MipNerf360).

\paragraph{Implementation.} We implement our method on top of the 3DGS-MCMC version from the \texttt{gsplat}~\cite{ye2025gsplat} framework. We adopt delentropy as the complexity functional since it attained the highest average rank in the pilot study (Tab.~\ref{tab:correlation}), although our method is agnostic to this choice and admits any complexity functional. On a single NVIDIA H100, reconstructing a typical scene with $\sim$2M primitives completes in 15 minutes. All hyperparameters are listed in Tab.~\ref{tab:hparams}.

\paragraph{Training details.}
\label{supp:training_details}
We train every scene for 30k steps. Each step renders the training view twice, once at full capacity and once at a budget $\tau\sim\mathcal{U}(0,1)$ sampled anew at every step, so that a single optimization covers the whole LoD spectrum. During the first $500$ steps we render at full capacity only, so that the full-capacity solution forms before any prefix is optimized. The total objective averages the two reconstruction terms with the variance term of Eq.~\ref{eq:var}, and adds the regularizers of 3DGS-MCMC:
\begin{equation}
  \mathcal{L} = \tfrac{1}{2}\bigl(\underbrace{\mathcal{L}_{\text{rec}}(\hat I_i^{\text{full}}, I_i)}_{\text{full capacity}}
  + \underbrace{\mathcal{L}_{\text{rec}}(\hat I_i^{\tau}, I_i)}_{\text{budget }\tau}
  + \underbrace{\lambda_{\text{var}}\,\mathcal{L}_{\text{var}}}_{\text{uniform error}}\bigr)
  + \underbrace{\lambda_{\alpha}\,\bar{\alpha}}_{\text{opacity}} + \underbrace{\lambda_{\sigma}\,\bar{\sigma}}_{\text{scale}},
  \label{eq:total}
\end{equation}
where $\mathcal{L}_{\text{rec}}$ is the L1+D-SSIM loss of 3DGS, and $\bar{\alpha}$ and $\bar{\sigma}$ are the mean opacity and mean scale of the primitives, weighted by $\lambda_{\alpha}$ and $\lambda_{\sigma}$ (Tab.~\ref{tab:hparams}). A newly added or relocated Gaussian has no complexity estimate until a training view observes it. We therefore set its complexity to the minimum, so relocation ranks it by opacity alone until the update of Eq.~\ref{eq:complexity_update} assigns it a value.

\begin{table}[h]
\centering
\label{tab:hyperparameters}
\begin{tabular}{l c}
\toprule
\textbf{Hyperparameter} & \textbf{Value} \\
\midrule
Spatial-variance loss weight \(\lambda_{\mathrm{var}}\) & \(2.0\) \\
Importance-score weight \(\lambda\) & \(2.0\) \\
Complexity EMA rate \(\eta\) & \(0.8\) \\
Training LoD sampling distribution & \(\tau \sim \mathcal{U}(0,100\%)\) \\
Dropout rate on normalized complexity \(\tilde{C}\) & \(0.2\) \\
Training steps & 30000 \\
\bottomrule
\end{tabular}
\vspace{0.25em}
\caption{\textbf{Hyperparameters used in our method.}}
\label{tab:hparams}
\end{table}
\vspace{-1.7em}

\section{Experimental evaluation.}
\label{supp:experimental_evaluation}

\subsection{Setup} We follow the evaluation protocol of 3DGS-MCMC~\cite{kheradmand20243dgsmcmc} and report results on MipNerf360~\cite{barron2022mipnerf} with 7 scenes, Tanks and Temples~\cite{knapitsch2017tandt} with 2 scenes, and Deep Blending (DB)~\cite{hedman2018deepblending} with 2 scenes, together covering a wide range of indoor and outdoor capture conditions. We compare against five baselines spanning the design space of LoD for Gaussian Splatting. Two of them yield a prefix renderable ordering directly comparable to ours: Matryoshka Gaussian Splatting (MGS)~\cite{guo2026matryoshka}, our closest competitor, and 3DGS-MCMC~\cite{kheradmand20243dgsmcmc}, which we extend with an opacity based ranking. Other two operate only at restricted points of the LoD spectrum: FlexGS~\cite{liu2025flexgs}, a budget conditioned continuous LoD method, and Octree-GS~\cite{ren2025octreegs}, a discrete LoD method. The last, LightGaussian~\cite{fan2024lightgaussian}, is a pruning method, included to compare our ordering against post hoc decimation at equal capacity. A detailed description of each baseline and its adaptation to our protocol is provided in Sec.~\ref{sec:supp_baselines}. All models are trained under identical conditions on NVIDIA H100 GPUs. For fair comparison we fix the total primitive budget across 3DGS-MCMC, MGS, and our method, while FlexGS and Octree-GS are reported at their native counts. We report PSNR~\cite{haskell1997psnr}, SSIM~\cite{wang2004ssim}, and LPIPS~\cite{zhang2018lpips}, which capture the complementary axes of pixel fidelity, structural preservation, and perceptual similarity.

\subsection{Baselines}
\label{sec:supp_baselines}
\paragraph{3DGS-MCMC~\cite{kheradmand20243dgsmcmc}.}
We use the \texttt{gsplat} reimplementation with default hyperparameters. Since 3DGS-MCMC is neither LOD-aware at training time nor designed to support budget-conditioned inference, we evaluate it under LOD by applying the simplest possible reduction strategy: at each target budget we retain the top-$k$ primitives ranked by opacity and discard the rest. This naive trimming establishes the baseline behaviour of an opacity-only ranking signal absent any LOD-aware training.
\vspace{-0.5em}
\paragraph{Matryoshka-GS~\cite{guo2026matryoshka}.}
We use the authors original implementation. One important difference from their reported setup affects direct comparability: the original paper trains all scenes with a flat 5M-primitive budget for 50k steps. To align with our protocol and with the 3DGS-MCMC budget convention, we instead train MGS for 30k steps and adapt the primitive budget per scene to match the 3DGS allocation. All other hyperparameters follow the original implementation. Notably by doing, we observed that the stability has improved over the DB dataset over the official results.
\vspace{-0.5em}
\paragraph{FlexGS~\cite{liu2025flexgs}.}
We use the authors' original implementation and training hyperparameters without modification. FlexGS is a budget-conditioned continuous-LOD method that learns a differentiable selector mapping each primitive to a retention score for a target budget, trained in two stages: a first stage optimises the 3DGS representation, and a second stage trains the selector on top of the frozen primitives. This design trades global consistency for pointwise quality: FlexGS can outperform prefix-ranked methods at the specific budgets it is optimised for, but at the cost of (i) roughly doubled end-to-end training time, and (ii) the absence of a prefix-renderable ordering, since changing the operating LOD requires a fresh forward pass through the selector to recompute per-primitive scores rather than truncating a precomputed ranking.
\vspace{-0.5em}
\paragraph{Octree-GS~\cite{ren2025octreegs}.}
We use the authors' original implementation and training hyperparameters without modification. Octree-GS is a discrete-LOD method that organises primitives into a fixed octree hierarchy, exposing a small number of predefined operating points corresponding to traversal depths. This design yields stable quality at its native levels but cannot interpolate between them: budgets falling between two depths are unreachable, and the coarsest level is bounded below by the structure of the hierarchy itself.
\vspace{-0.5em}
\paragraph{LightGaussian~\cite{fan2024lightgaussian}.} We use the authors' original implementation. LightGaussian is a compression method rather than an LoD one, pruning primitives by a global volumetric importance measure, fine tuning the survivors, then compacting them through spherical harmonic distillation and vector quantisation. We retain only the pruning stage, as the others reduce storage without changing the primitive count that defines our budget axis. For fairness we initialise from the same 30k step 3DGS-MCMC checkpoints as the other baselines so every method starts from an identical primitive set. We omit the recovery fine tuning, since fine tuning at each budget yields a separate optimisation run and a stored model per operating point, which is what separates pruning from a single prefix renderable ordering.

\vspace{-0.4em}
\section{Additional experiments}
\label{supp:sec_additional_exp}
\vspace{-0.6em}
In all tables, best results are in \textbf{bold} and second best are \underline{underlined}.

\subsection{Additional qualitative results}
\label{sec:supp_qualitative}
\begin{figure}[h!]
    \centering
    \includegraphics[width=\linewidth]{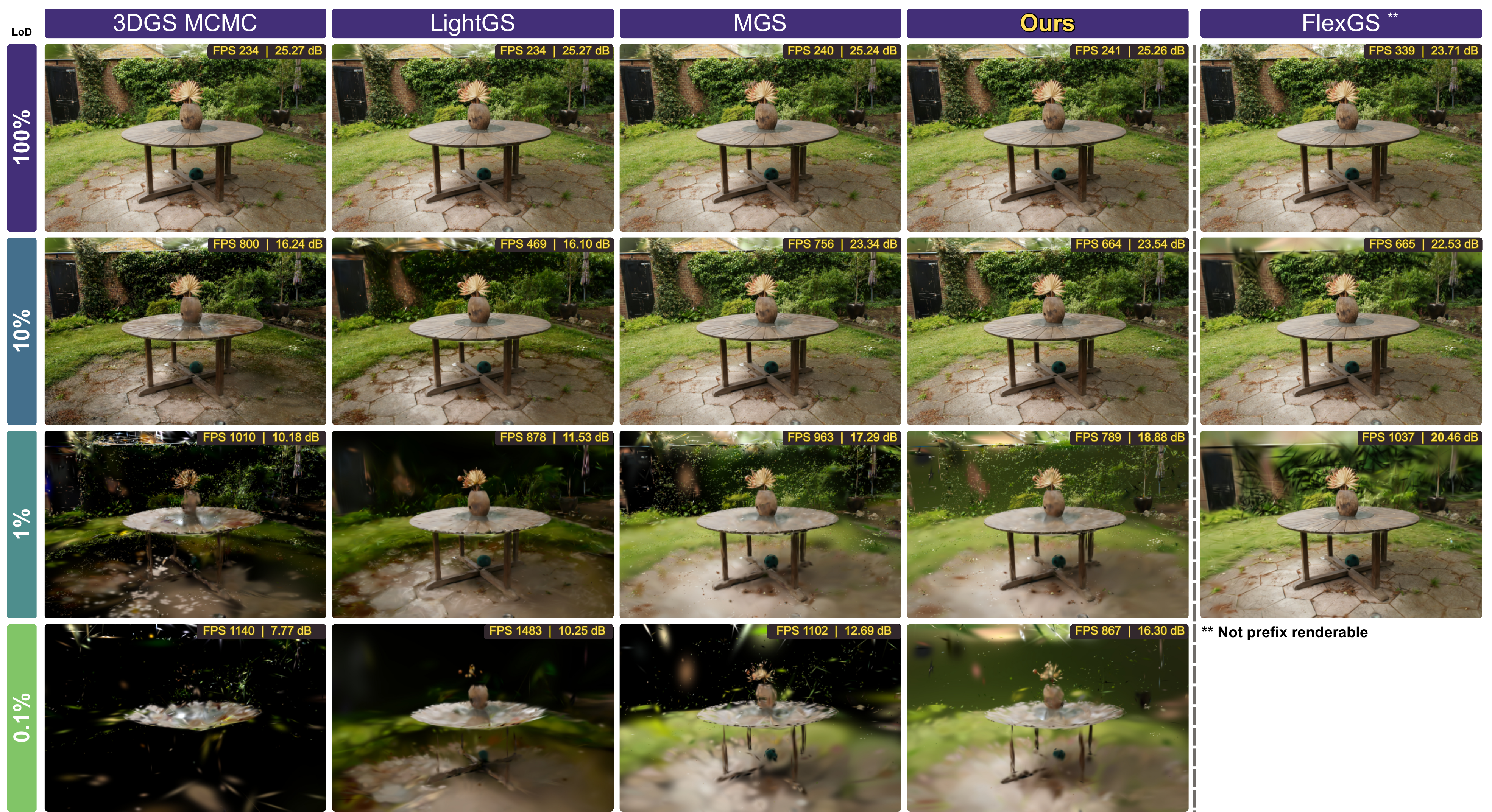}
    \caption{\textbf{Qualitative comparison on Garden.} Renders of the outdoor Garden scene from {MipNerf360}~\cite{barron2022mipnerf} at four budgets.}
    \label{fig:supp_garden}
\end{figure}

\begin{figure}[h!]
    \centering
    \includegraphics[width=\linewidth]{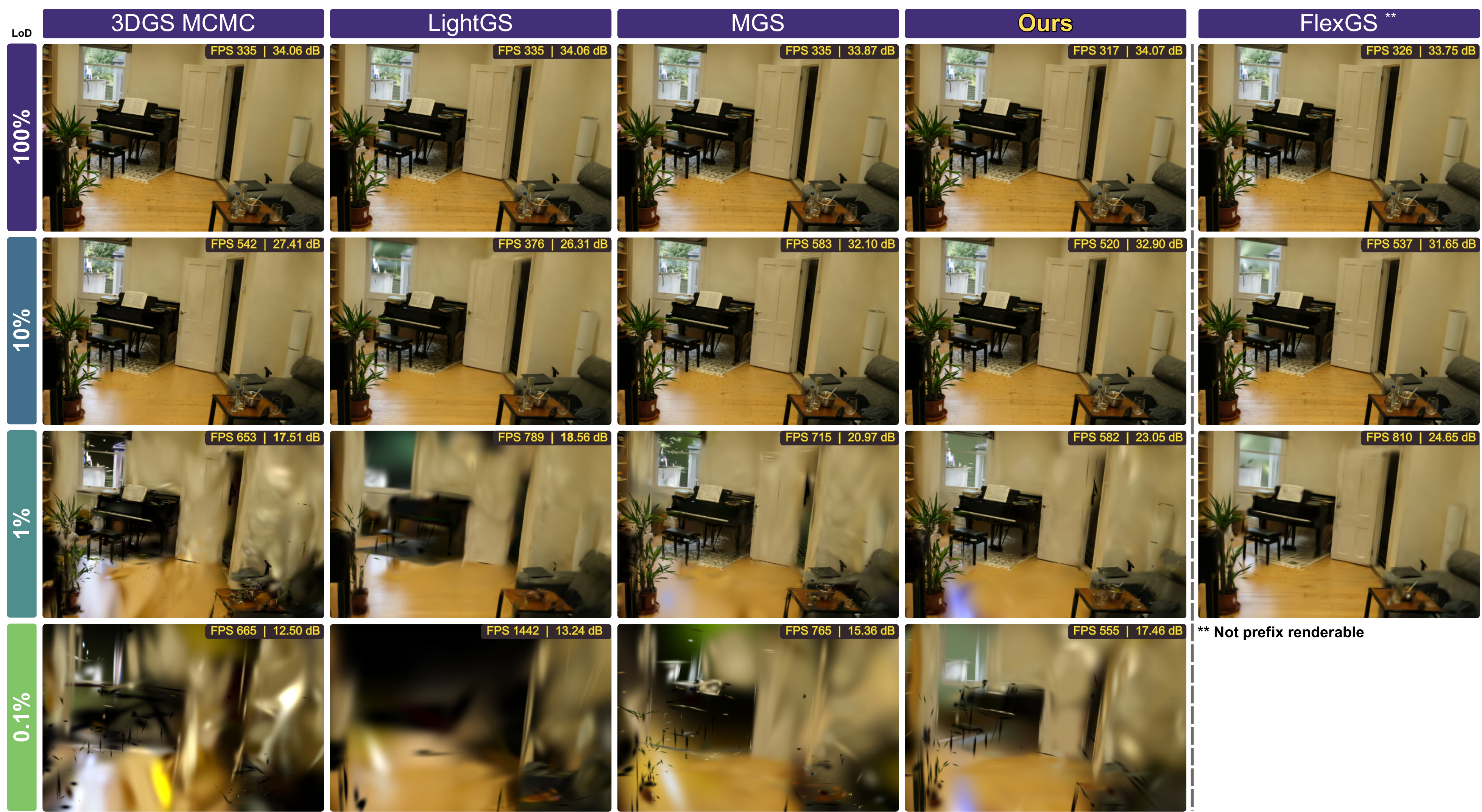}
    \caption{\textbf{Qualitative comparison on Room.} Renders of the indoor Room scene from {MipNerf360}~\cite{barron2022mipnerf} at four budgets.}
    \label{fig:supp_room}
\end{figure}

We complement Fig.~\ref{fig:visual_quality} with two more MipNerf360 dataset~\cite{barron2022mipnerf} scenes, the outdoor Garden (Fig.~\ref{fig:supp_garden}) and the indoor Room (Fig.~\ref{fig:supp_room}). Each row shows one budget and each column one method. Insets report the FPS measured over the entire scene and the PSNR of the presented image.

\subsection{Dropout ablation}
\label{supp:subsec_drout_ablation}
In this section, we investigate how the complexity dropout rate $\delta$ used during relocation (Eq.~\ref{eq:adc_proposal}) affects reconstruction quality. With this dropout, the complexity term $\tilde{C}_j$ of a candidate $g_j$ is set to zero with probability $\delta$, in which case $g_j$ is not removed from the candidate set but competes for relocation based only on its opacity. The rate $\delta$ therefore interpolates between two extremes: at $\delta = 0$, the complexity term is always applied and the relocation weights are fully determined by the complexity field, while at $\delta = 1$, the complexity term is always dropped and the rule reduces to the original opacity sampling $p(g_j) = \alpha_j$. We perform a value sweep $\delta \in \{0, 0.2, 0.4, 0.6, 0.8, 1\}$ on the outdoor Tanks and Temples~\cite{knapitsch2017tandt} and the indoor Deep Blending~\cite{hedman2018deepblending} datasets, and report the metrics at full capacity and the AUC over $[0.1\%, 100\%]$ (Tab.~\ref{tab:dropout}). Our default value $\delta = 0.2$ attains the best result on all twelve entries, tied with $\delta = 0.4$ only for SSIM at full capacity on Deep Blending. Compared to the original opacity sampling ($\delta = 1$), it improves the AUC PSNR by $0.14$\,dB on TandT ($24.02$ vs.\ $23.88$) and by $0.23$\,dB on Deep Blending ($29.51$ vs.\ $29.28$). Applying the complexity term at every step ($\delta = 0$) is not beneficial either, as it falls below opacity sampling on Deep Blending ($29.00$ vs.\ $29.28$ AUC PSNR). We hypothesize that without dropout, relocation repeatedly targets the same high complexity regions, whereas occasionally dropping the complexity term lets opacity sampling reach the remaining regions. Large intermediate rates ($\delta \in \{0.6, 0.8\}$) perform worst on both datasets. Although the differences are small relative to the noise of the metrics, the consistency of $\delta = 0.2$ across both datasets supports it as a robust default. We also observe more stable training with complexity dropout, especially on the indoor scenes of Deep Blending.

\begin{table}[t]
\centering
\resizebox{0.7\linewidth}{!}{%
\begin{tabular}{ll ccc ccc}
\toprule
& & \multicolumn{3}{c}{\textbf{100\% Capacity}} & \multicolumn{3}{c}{\textbf{AUC [0.1\%, 100\%]}} \\
\cmidrule(lr){3-5} \cmidrule(lr){6-8}
\textbf{Dataset} & $\delta$ & PSNR $\uparrow$ & SSIM $\uparrow$ & LPIPS $\downarrow$ & PSNR $\uparrow$ & SSIM $\uparrow$ & LPIPS $\downarrow$ \\
\midrule
\multirow{6}{*}{TandT~\cite{knapitsch2017tandt}} & 0 & \underline{24.52} & 0.856 & 0.153 & \underline{23.91} & 0.829 & 0.193 \\
& \cellcolor{gray!10}0.2 & \cellcolor{gray!10}\textbf{24.65} & \cellcolor{gray!10}\textbf{0.868} & \cellcolor{gray!10}\textbf{0.150} & \cellcolor{gray!10}\textbf{24.02} & \cellcolor{gray!10}\textbf{0.839} & \cellcolor{gray!10}\textbf{0.189} \\
& 0.4 & 24.46 & 0.852 & 0.152 & 23.86 & 0.828 & 0.192 \\
& 0.6 & 24.44 & 0.847 & 0.163 & 23.85 & 0.818 & 0.195 \\
& 0.8 & 24.39 & 0.841 & 0.171 & 23.80 & 0.811 & 0.199 \\
& 1 & 24.45 & \underline{0.865} & \underline{0.151} & 23.88 & \underline{0.837} & \underline{0.190} \\
\midrule
\multirow{6}{*}{DB~\cite{hedman2018deepblending}} & 0 & 29.39 & 0.905 & 0.246 & 29.00 & 0.896 & \underline{0.263} \\
& \cellcolor{gray!10}0.2 & \cellcolor{gray!10}\textbf{29.92} & \cellcolor{gray!10}\textbf{0.911} & \cellcolor{gray!10}\textbf{0.240} & \cellcolor{gray!10}\textbf{29.51} & \cellcolor{gray!10}\textbf{0.902} & \cellcolor{gray!10}\textbf{0.254} \\
& 0.4 & 29.63 & \textbf{0.911} & \underline{0.241} & 29.23 & \underline{0.900} & 0.264 \\
& 0.6 & 29.21 & \underline{0.909} & 0.252 & 28.94 & 0.882 & 0.269 \\
& 0.8 & 29.19 & 0.899 & 0.253 & 28.90 & 0.885 & 0.271 \\
& 1 & \underline{29.75} & 0.903 & 0.247 & \underline{29.28} & 0.899 & \underline{0.263} \\
\bottomrule
\end{tabular}%
}
\vspace{0.3em}
\caption{\textbf{Complexity dropout rate.} Sweep of the dropout rate $\delta$ on TandT and DB. $\delta = 0$ always applies the complexity term and $\delta = 1$ reduces to opacity sampling. Our default is shaded.}
\label{tab:dropout}
\vspace{-0.5em}
\end{table}

\subsection{Rendering Speed}
\label{supp:subsec_rendering_speed}

We investigate the rendering speed of our method against the baselines at four levels of detail ($100\%$, $10\%$, $1\%$, $0.1\%$). We split MipNeRF360~\cite{barron2022mipnerf} into indoor scenes (Bonsai, Counter, Kitchen, Room) and outdoor scenes (Bicycle, Garden, Stump) and report the average frame rate on a single NVIDIA H100 (Tab.~\ref{tab:supp_fps}). All prefix methods share an identical budget, so their primitive counts match at every level of detail and speed differences stem only from the size of the retained primitives, whereas FlexGS~\cite{liu2025flexgs} operates with a smaller budget and is not directly comparable. At full capacity, our method is the fastest prefix method, tied with MGS indoors ($302$ FPS) and ahead outdoors ($198$ vs.\ $193$ FPS). As the budget shrinks, every method accelerates, but ours does so less and saturates below $1\%$, reaching $463$ FPS indoors and $874$ FPS outdoors at $0.1\%$, against $829$ and $1{,}151$ FPS for MGS. We attribute this to our importance score (Eq.~\ref{eq:score}), which retains large primitives so that simple regions remain covered. A large primitive spans more tiles and thus costs more per frame, whereas MGS favors compact, high opacity primitives that render faster but cover less of the scene, at lower quality (Fig.~\ref{fig:decay}). Our method is thus slower than MGS at low budgets, a cost we accept in exchange for coverage and better quality, while at equal quality it requires fewer primitives (Fig.~\ref{fig:ttq}). Reducing the per frame cost of large primitives at the lowest budgets, for instance by bounding their screen footprint, is a natural direction for future work.

\begin{table}[t]
\centering
\setlength{\tabcolsep}{5pt}
\resizebox{0.7\linewidth}{!}{%
\begin{tabular}{l cccc c cccc}
\toprule
 & \multicolumn{4}{c}{Indoor} & & \multicolumn{4}{c}{Outdoor} \\
\cmidrule{2-5} \cmidrule{7-10}
Method & $100\%$ & $10\%$ & $1\%$ & $0.1\%$ & & $100\%$ & $10\%$ & $1\%$ & $0.1\%$ \\
\midrule
FlexGS$^{*}$~\cite{liu2025flexgs}            & 339 & 578 & 833  & --   & & 351 & 609 & 1021 & --   \\
\arrayrulecolor{black!30}\midrule\arrayrulecolor{black}
LightGaussian~\cite{fan2024lightgaussian}   & \underline{300} & 379 & \underline{697} & \textbf{1101} & & 189 & 435 & 855 & \textbf{1364} \\
3DGS MCMC~\cite{kheradmand20243dgsmcmc}     & \underline{300} & \underline{568} & 688 & 732 & & 189 & \textbf{805} & \textbf{1093} & \underline{1177} \\
MGS~\cite{guo2026matryoshka}                & \textbf{302} & \textbf{595} & \textbf{731} & \underline{829} & & \underline{193} & \underline{769} & \underline{1050} & 1151 \\
\rowcolor{black!8}
\textbf{Ours}                               & \textbf{302} & 450 & 475 & 463 & & \textbf{198} & 649 & 833 & 874 \\
\bottomrule
\end{tabular}
}

\vspace{0.2em}
{\footnotesize $^{*}$FlexGS operates with a smaller primitive budget, hence the faster rendering speed.}
\vspace{0.3em}
\caption{\textbf{Rendering speed.} Average frame rate (FPS $\uparrow$) on MipNerf360 at four levels of detail. Indoor scenes are rendered at half resolution and outdoor scenes at quarter resolution. FlexGS is excluded from the ranking.}
\label{tab:supp_fps}
\end{table}

\subsection{Degradation Across Complexity Levels}
\label{supp:subsec_degradation_across_complexity_levels}

\begin{figure}[t]
\centering
\begin{minipage}[t]{0.45\linewidth}
  \centering
  \includegraphics[width=\linewidth]{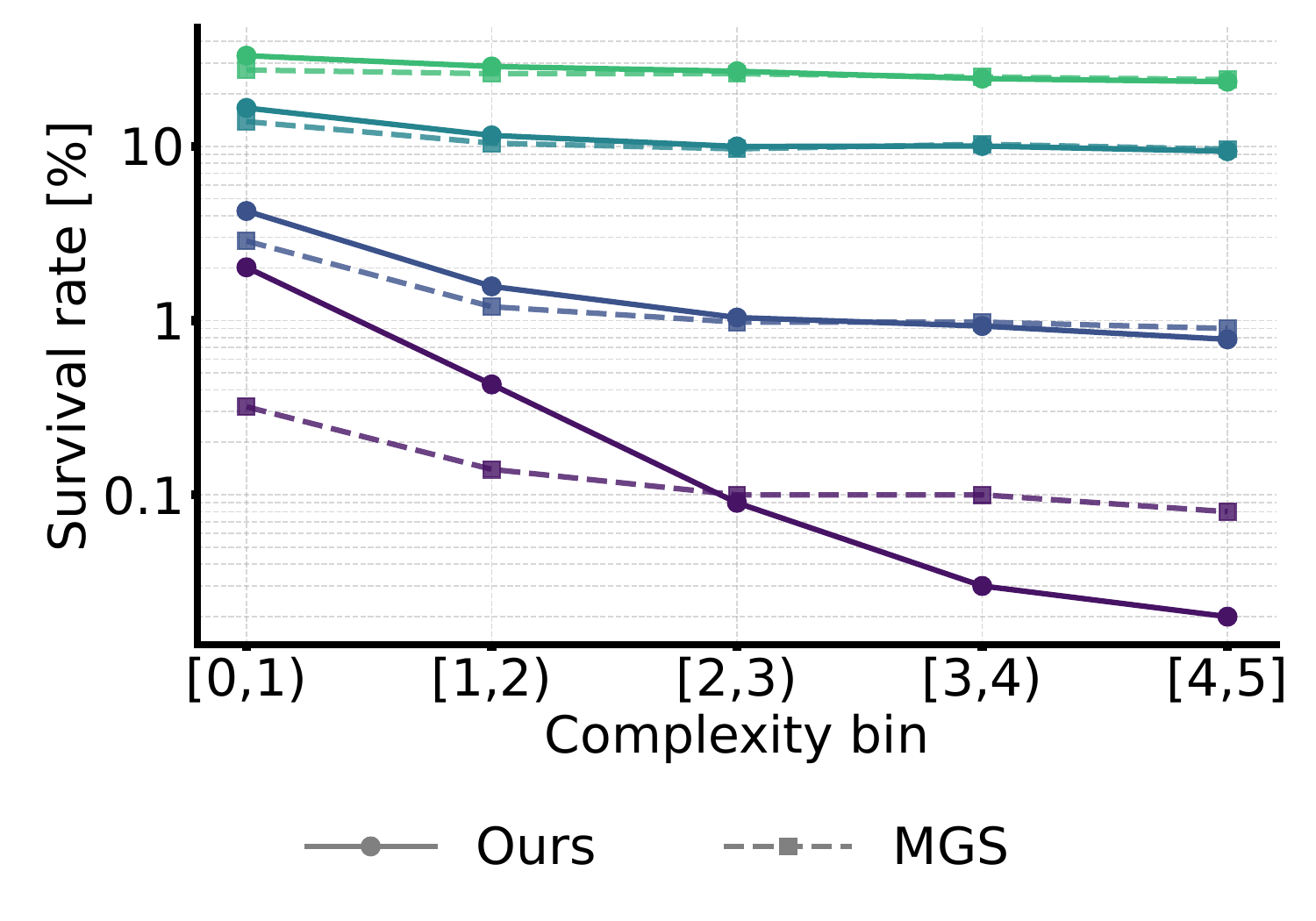}
  \caption{\textbf{Survival rate.} Percentage of the Gaussians of each complexity bin at full capacity that are still retained at a given budget (log scale). Our method keeps more Gaussians in simple bins and fewer in complex bins than MGS.}
  \label{fig:survival_rate}
\end{minipage}\hfill
\begin{minipage}[t]{0.45\linewidth}
  \centering
  \includegraphics[width=\linewidth]{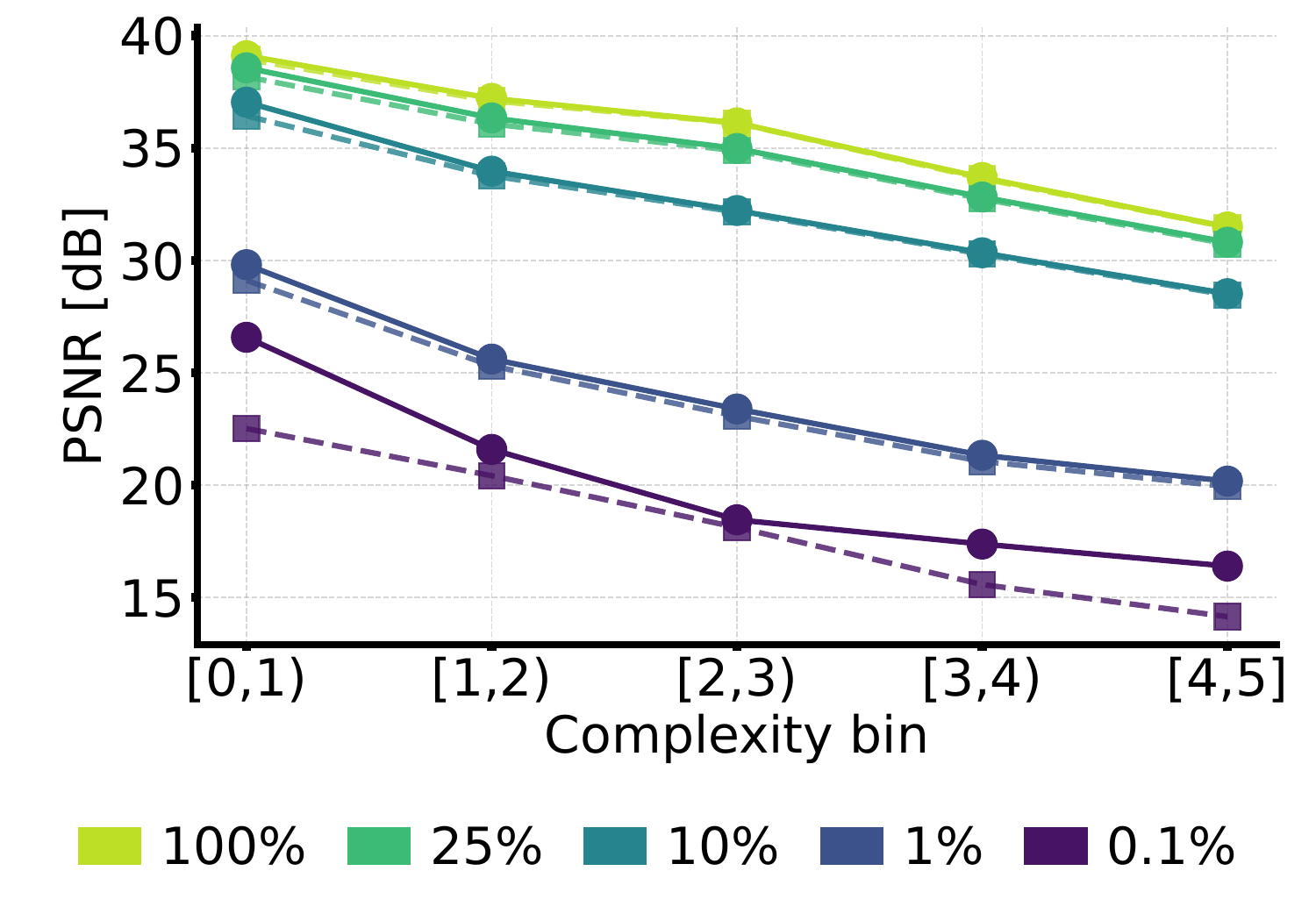}
  \caption{\textbf{Quality.} Average PSNR of the Gaussians in each complexity bin, computed from the per pixel error maps projected onto the Gaussians. Our method has the largest gains at the lowest capacity.}
  \label{fig:bin_psnr}
\end{minipage}
\vspace{1.0em}
\end{figure}

In this section, we investigate whether the importance score (Eq.~\ref{eq:score}) undoes the capacity that complexity guided densification (Eq.~\ref{eq:adc_proposal}) adds to complex regions. Since the score reduces to opacity in complex regions, primitives there are ranked with less spatial support and are therefore discarded earlier as the budget shrinks. To quantify this effect, we project the image complexity maps onto the Gaussians with the same aggregation as in our pilot study (Sec.~\ref{sec:pilot}), group the Gaussians into five complexity bins, and compare our method with MGS~\cite{guo2026matryoshka} through three statistics per bin as follows. (i)~\emph{Allocation}: the number of Gaussians in each bin at full capacity, which shows where densification places the primitives (Tab.~\ref{tab:bin_allocation}). (ii)~\emph{Survival rate}: for each bin, the percentage of its full capacity Gaussians that are still present at a given budget, \eg, a survival rate of $10\%$ means that one in ten Gaussians of that bin is kept (Fig.~\ref{fig:survival_rate}). (iii)~\emph{Quality}: we project the per pixel error maps onto the Gaussians with the same aggregation and report the average error of each bin as~PSNR~(Fig.~\ref{fig:bin_psnr}). At full capacity, densification shifts the population towards complex regions, with $472.9$k Gaussians in the most complex bin $[4,5]$ against $460.5$k for MGS, and $21.1$k against $22.9$k in the simplest bin $[0,1)$. As the budget shrinks, survival drops with complexity, as intended by the score: at $25\%$ budget, our method retains $33.08\%$ of the Gaussians in $[0,1)$ against $23.47\%$ in $[4,5]$, and at $0.1\%$ the ratio grows to $2.02\%$ against $0.02\%$. In complex bins, our method thus keeps fewer Gaussians than MGS at low budgets ($0.02\%$ vs.\ $0.08\%$ in $[4,5]$ at $0.1\%$). This reduced survival does not translate into lower quality, as our method attains a higher PSNR than MGS in every bin and at every budget, including the most complex bin, where the gap grows from $+0.02$\,dB at $100\%$ to $+2.25$\,dB at $0.1\%$ ($16.40$ vs.\ $14.15$). These results suggest that the Gaussians retained in complex regions are better placed by densification, so that the capacity removed by the score is compensated rather than reintroduced as a failure mode.

\begin{table}[t]
\centering
\begin{tabular}{ll ccccc}
\toprule
\textbf{Method} & \textbf{Statistic} & $[0,1)$ & $[1,2)$ & $[2,3)$ & $[3,4)$ & $[4,5]$ \\
\midrule
\multirow{2}{*}{MGS~\cite{guo2026matryoshka}} & Count & 22.9k & 110.0k & 240.1k & 466.5k & 460.5k \\
& Share & 1.8\% & 8.5\% & 18.5\% & 35.8\% & 35.4\% \\
\midrule
\multirow{2}{*}{Ours} & Count & 21.1k & 100.7k & 233.8k & 471.5k & 472.9k \\
& Share & 1.6\% & 7.7\% & 18.0\% & 36.3\% & 36.4\% \\
\bottomrule
\end{tabular}
\vspace{0.3em}
\caption{\textbf{Full capacity allocation.} Number and share of Gaussians per complexity bin at a budget of 1300k. Our densification shifts Gaussians towards the most complex bins.}
\vspace{-0.7em}
\label{tab:bin_allocation}
\end{table}
\vspace{-0.5em}

\subsection{Extended ablation}
\label{sec:supp_ablation}

We complement the aggregate results of Sec.~\ref{sec:ablation} with a breakdown at individual LoD budgets on MipNeRF360. Table~\ref{tab:supp_ablation_full} reports all component combinations at $100\%$, $10\%$, $1\%$, and $0.1\%$ capacity, alongside the AUC values reported in the main paper. Among the individual components, importance scoring produces the strongest improvements across all three metrics at every reduced budget, increasing PSNR at $0.1\%$ from $14.59$ to $15.94$\,dB. Densification also improves low-budget quality, reaching $15.68$\,dB at $0.1\%$, while the variance loss alone provides smaller gains. Combining densification and scoring further raises PSNR to $16.08$\,dB at this budget, and adding the variance loss brings it to $16.18$\,dB. The complete method achieves the best results across all metrics and evaluated budgets. Its PSNR gain over the baseline increases from $0.25$\,dB at full capacity to $0.80$, $0.93$, and $1.59$\,dB at $10\%$, $1\%$, and $0.1\%$, respectively, showing that the improvements are most pronounced under aggressive budget reduction.

\begin{table}[t]
\centering
\resizebox{\textwidth}{!}{
\begin{tabular}{l ccc ccc ccc ccc ccc ccc}
\toprule
& \multicolumn{3}{c}{\textbf{Components}} & \multicolumn{3}{c}{\textbf{AUC [0.1\%, 100\%]}} & \multicolumn{3}{c}{\textbf{100\% Capacity}} & \multicolumn{3}{c}{\textbf{10\% Capacity}} & \multicolumn{3}{c}{\textbf{1\% Capacity}} & \multicolumn{3}{c}{\textbf{0.1\% Capacity}} \\
\cmidrule(lr){2-4} \cmidrule(lr){5-7} \cmidrule(lr){8-10} \cmidrule(lr){11-13} \cmidrule(lr){14-16} \cmidrule(lr){17-19}
& Dens. & Score & Var. & PSNR $\uparrow$ & SSIM $\uparrow$ & LPIPS $\downarrow$ & PSNR $\uparrow$ & SSIM $\uparrow$ & LPIPS $\downarrow$ & PSNR $\uparrow$ & SSIM $\uparrow$ & LPIPS $\downarrow$ & PSNR $\uparrow$ & SSIM $\uparrow$ & LPIPS $\downarrow$ & PSNR $\uparrow$ & SSIM $\uparrow$ & LPIPS $\downarrow$ \\
\midrule
None & \xmark & \xmark & \xmark & 28.74 & 0.855 & 0.196 & 29.55 & 0.863 & 0.164 & 27.02 & 0.794 & 0.275 & 20.12 & 0.543 & 0.540 & 14.59 & 0.393 & 0.623 \\
\midrule
\multirow{3}{*}{Single} & \cmark & \xmark & \xmark & 28.81 & 0.859 & 0.190 & \underline{29.79} & \underline{0.889} & \underline{0.151} & 27.53 & 0.805 & 0.263 & 20.76 & 0.560 & 0.511 & 15.68 & 0.424 & 0.616 \\
& \xmark & \cmark & \xmark & 28.94 & 0.861 & 0.188 & 29.62 & 0.882 & 0.159 & 27.69 & 0.818 & \underline{0.259} & 20.95 & 0.564 & 0.511 & 15.94 & \underline{0.457} & \underline{0.615} \\
& \xmark & \xmark & \cmark & 28.79 & 0.855 & 0.190 & 29.59 & 0.870 & 0.160 & 27.44 & 0.804 & 0.268 & 20.39 & 0.553 & 0.520 & 15.10 & 0.412 & 0.621 \\
\midrule
\multirow{3}{*}{Pair} & \cmark & \cmark & \xmark & \underline{28.99} & \underline{0.862} & \underline{0.187} & 29.75 & 0.885 & 0.154 & \underline{27.75} & \underline{0.819} & \underline{0.259} & \underline{20.99} & \underline{0.565} & \underline{0.510} & \underline{16.08} & \underline{0.457} & \underline{0.615} \\
& \cmark & \xmark & \cmark & 28.75 & 0.850 & 0.197 & 29.70 & 0.877 & 0.159 & 27.61 & 0.812 & 0.262 & 20.70 & 0.556 & 0.514 & 15.22 & 0.415 & 0.620 \\
& \xmark & \cmark & \cmark & 28.95 & 0.860 & 0.189 & 29.71 & 0.880 & 0.155 & 27.74 & 0.816 & 0.260 & 20.77 & 0.559 & 0.512 & 15.96 & 0.456 & \underline{0.615} \\
\midrule
\rowcolor{gray!10} All & \cmark & \cmark & \cmark & \textbf{29.09} & \textbf{0.867} & \textbf{0.184} & \textbf{29.80} & \textbf{0.893} & \textbf{0.150} & \textbf{27.82} & \textbf{0.822} & \textbf{0.257} & \textbf{21.05} & \textbf{0.566} & \textbf{0.508} & \textbf{16.18} & \textbf{0.458} & \textbf{0.612} \\
\bottomrule
\end{tabular}
}\vspace{0.3em}
\caption{\textbf{Full ablation.} All combinations of complexity aware densification (Dens.), importance scoring (Score), and the spatial variance loss (Var.) on MipNeRF360.}
\vspace{-1.0em}
\label{tab:supp_ablation_full}
\end{table}

\subsection{Coverage Across LoDs}
\label{sec:supp_coverage}
\begin{table}[t]
\centering
\begin{subtable}[t]{0.48\linewidth}
\centering
\begin{tabular}{l cc}
\toprule
\textbf{LoD} & \textbf{MGS}\cite{guo2026matryoshka} \# holes $\downarrow$ & \textbf{Ours} \# holes $\downarrow$ \\
\midrule
100\% & 0 & 0 \\
10\% & 0 & 0 \\
1\% & 2,615 & \textbf{2,420} \\
0.1\% & 83,341 & \textbf{20,765} \\
\bottomrule
\end{tabular}
\caption{Bonsai}
\label{tab:supp_coverage_bonsai}
\end{subtable}\hfill
\begin{subtable}[t]{0.48\linewidth}
\centering
\begin{tabular}{l cc}
\toprule
\textbf{LoD} & \textbf{MGS}\cite{guo2026matryoshka} \# holes $\downarrow$ & \textbf{Ours} \# holes $\downarrow$ \\
\midrule
100\% & 0 & 0 \\
10\% & 0 & 0 \\
1\% & 11,884 & \textbf{2,204} \\
0.1\% & 116,920 & \textbf{36,009} \\
\bottomrule
\end{tabular}
\caption{Bicycle}
\label{tab:supp_coverage_bicycle}
\end{subtable}
\vspace{0.3em}
\caption{\textbf{Coverage across LoDs.} Number of "hole" pixels, \ie, pixels whose accumulated opacity is below $0.05$, on the test set of the indoor Bonsai and outdoor Bicycle scenes\cite{barron2022mipnerf}. Lower is better.}
\label{tab:supp_coverage}
\end{table}

In this experiment, we investigate whether the primitives retained at low budgets still cover the whole scene. A reduced budget can degrade a rendering in two ways: the scene may be represented at a coarser resolution, or parts of it may no longer be covered by any primitive, leaving "holes". We further define such holes as a pixel whose accumulated opacity is $\alpha < 0.05$ and count such pixels over all test views at 4 LoDs (100, 10, 1, 0.1). We compare against MGS~\cite{guo2026matryoshka}, our closest prefix renderable competitor, on two MipNerf360 scenes~\cite{barron2022mipnerf}: the indoor Bonsai and the outdoor Bicycle. Tab.~\ref{tab:supp_coverage} shows that coverage is preserved by both methods at high budgets and that the differences emerge as the budget shrinks. At $100\%$ and $10\%$ capacity, neither method leaves any hole. At $1\%$ capacity, the two methods are close on Bonsai ($2{,}420$ vs.\ $2{,}615$ hole pixels), whereas on Bicycle MGS already leaves $5.4\times$ more holes than ours ($11{,}884$ vs.\ $2{,}204$). At $0.1\%$ capacity, our method leaves fewer holes on both scenes, with MGS producing $4.0\times$ more on Bonsai ($83{,}341$ vs.\ $20{,}765$) and $3.2\times$ more on Bicycle ($116{,}920$ vs.\ $36{,}009$). We attribute this difference to our importance score~(Eq.~\ref{eq:score}), which retains large primitives in simple regions, so that areas such as the background remain covered by a few large Gaussians as the budget shrinks. The same mechanism explains why the gap appears earlier on Bicycle, whose large and simple outdoor background is the first to be removed by an ordering that does not account for coverage.

\FloatBarrier
\section{Inter and Intra LOD Variance Measure}
\label{sec:lod_variance_measure}

We formally define here the inter and intra LoD variance measurement. Let \(\Delta_t^m(p)\) denote the direction-corrected quality degradation of patch \(p\) between two consecutive LoDs \(\tau_t \rightarrow \tau_{t+1}\), the inter LoD variance writes:

\begin{equation}
V_{\mathrm{inter}}^m
=
\frac{1}{|\mathcal P|}
\sum_{p\in\mathcal P}
\operatorname{Var}_{t}\!\left[\Delta_t^m(p)\right]\,,
\end{equation}

which measures whether the same patch degrades smoothly across consecutive LoD transitions. Similarly, the intra LoD variance measurement writes:

\begin{equation}
V_{\mathrm{intra}}^m
=
\frac{1}{T-1}
\sum_{t=1}^{T-1}
\operatorname{Var}_{p\in\mathcal P}\!\left[\Delta_t^m(p)\right]\,,
\end{equation}

and measures whether one LoD transition degrades the image uniformly across patches.